%% file: main.tex
\documentclass[lettersize,journal,compsoc]{IEEEtran}

\ifCLASSOPTIONcompsoc
  \usepackage[nocompress]{cite}
\else
  \usepackage{cite}
\fi

\usepackage{amsfonts}
\usepackage{amsmath}
\usepackage{algorithmic}
\usepackage{algorithm}
\usepackage{array}
\usepackage{booktabs}
\usepackage[caption=false,font=normalsize,labelfont=sf,textfont=sf]{subfig}
\usepackage[breaklinks=true,colorlinks,citecolor=blue,linkcolor=red]{hyperref}
\usepackage{gensymb}
\usepackage{graphicx}
\usepackage{multirow}
\usepackage{tabularx}
\usepackage{pifont}
\usepackage{url}

\usepackage{tikz}
\usepackage{pgf, pgfsys, pgffor}
\usepackage{pgfplotstable}
\pgfplotsset{compat=1.15}
\usepackage{xspace}
\usepackage[table]{xcolor}
\definecolor{rred}{RGB}{245, 152, 153}
\definecolor{oorange}{RGB}{253, 205, 154}
\definecolor{yyellow}{RGB}{255, 255, 153}

\ifCLASSOPTIONcompsoc
  \usepackage[nocompress]{cite}
\else
  \usepackage{cite}
\fi

\usepackage{textcomp}
\usepackage{stfloats}
\usepackage{verbatim}
\usepackage{utfsym}
\usepackage{cleveref}
\usepackage[breaklinks=true,colorlinks,citecolor=blue,linkcolor=red]{hyperref}

\def\eg{\textit{e.g.}}

\def\methodname{ArtHOI\xspace}
\definecolor{fakerefcolor}{rgb}{0.21,0.49,0.74}

\begin{document}

\title{\methodname: Articulated Human-Object Interaction Synthesis by 4D Reconstruction\\ from Video Priors}

\author{
    Zihao~Huang,
    Tianqi~Liu,
    Zhaoxi~Chen,
    Shaocong~Xu,
    Saining~Zhang, \\
    Lixing~Xiao,
    Zhiguo~Cao,
    Wei~Li,
    Hao~Zhao,
    and Ziwei~Liu
    \IEEEcompsocitemizethanks{\IEEEcompsocthanksitem Zihao Huang and Tianqi Liu are with the School of AIA, Huazhong University of Science and Technology, Wuhan 430074, China; the S-Lab, Nanyang Technological University (NTU), Singapore 639798; and the Beijing Academy of Artificial Intelligence (BAAI), Beijing 100083, China. \\
    Zhaoxi Chen, Wei Li, and Ziwei Liu are with the S-Lab, Nanyang Technological University (NTU), Singapore 639798. \\
    Shaocong Xu is with the Beijing Academy of Artificial Intelligence (BAAI), Beijing 100083, China. \\
    Saining Zhang is with the Nanyang Technological University (NTU), Singapore 639798, and also with the Beijing Academy of Artificial Intelligence (BAAI), Beijing 100083, China. \\
    Lixing Xiao is with Zhejiang University (ZJU), Hangzhou 310058, China. \\
    Zhiguo Cao is with the School of AIA, Huazhong University of Science and Technology, Wuhan 430074, China. \\
    Hao Zhao is with the Institute for AI Industry Research (AIR), Tsinghua University (THU), Beijing 100084, China, and also with the Beijing Academy of Artificial Intelligence (BAAI), Beijing 100083, China.}
    \IEEEcompsocitemizethanks{\IEEEcompsocthanksitem Corresponding authors: Hao Zhao and Ziwei Liu. Emails: \texttt{zhaohao@air.tsinghua.edu.cn}, \texttt{zwliu.hust@gmail.com}.}
    \thanks{}
}

\IEEEpubid{}
\IEEEtitleabstractindextext{
    \input{sections/0_abstract}
}
\maketitle
\input{sections/1_introduction}
\input{sections/2_related_work}
\input{sections/3_method}

\input{sections/4_experiments}

\input{sections/5_conclusion}

\ifCLASSOPTIONcompsoc
  \section*{Acknowledgments}
\else
  \section*{Acknowledgment}
\fi

This research is supported by cash and in-kind funding from NTU S-Lab and industry partner(s). This study is also supported by the Ministry of Education, Singapore, under its MOE AcRF Tier 2 (MOE-T2EP20221-0012, MOE-T2EP20223-0002).

\clearpage
\bibliographystyle{IEEEtran}
\bibliography{main}

\end{document}

%% file: sections/0_abstract.tex
\begin{abstract}
    Synthesizing physically plausible articulated human-object interactions (HOI) without 3D/4D supervision remains a fundamental challenge.
    While recent zero-shot approaches leverage video diffusion models to synthesize human-object interactions, they are largely confined to rigid-object manipulation and lack explicit 4D geometric reasoning. 
    To bridge this gap, we formulate articulated HOI synthesis as a \textbf{4D reconstruction problem from monocular video priors}: given only a video generated by a diffusion model, we reconstruct a full 4D articulated scene without any 3D supervision.
    This reconstruction-based approach treats the generated 2D video as supervision for an inverse rendering problem, recovering geometrically consistent and physically plausible 4D scenes that naturally respect contact, articulation, and temporal coherence.
    We introduce \textbf{\methodname}, the first zero-shot framework for articulated human-object interaction synthesis via 4D reconstruction from video priors. Our key designs are: \textbf{1)} \textit{Flow-based part segmentation}: leveraging optical flow as a geometric cue to disentangle dynamic from static regions in monocular video; \textbf{2)} \textit{Decoupled reconstruction pipeline}: joint optimization of human motion and object articulation is unstable under monocular ambiguity, so we first recover object articulation, then synthesize human motion conditioned on the reconstructed object states.
    \methodname bridges video-based generation and geometry-aware reconstruction, producing interactions that are both semantically aligned and physically grounded.
    Across diverse articulated scenes (\eg, opening fridges, cabinets, microwaves), \methodname significantly outperforms prior methods in contact accuracy, penetration reduction, and articulation fidelity, extending zero-shot interaction synthesis beyond rigid manipulation through reconstruction-informed synthesis.
    Project page is available at \url{https://arthoi.github.io/}.
\end{abstract}
\begin{IEEEkeywords}
Articulated Human-object Interaction, 4D Reconstruction, Motion Synthesis, Modeling from Video
\end{IEEEkeywords}

%% file: sections/1_introduction.tex
\section{Introduction}
\label{sec:intro}

\begin{figure*}[t]
    \centering
    \includegraphics[width=0.99\linewidth]{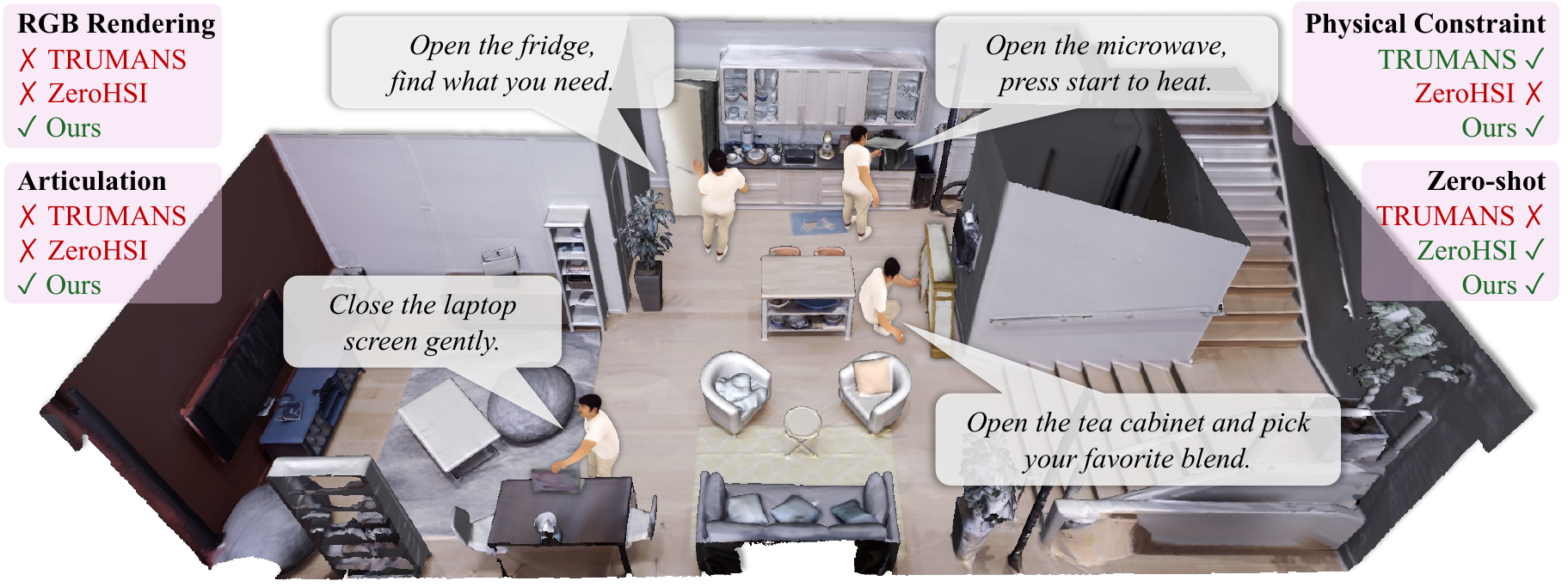}
    \caption{\methodname recovers zero-shot articulated human-object scene geometry and dynamics from monocular video priors without 3D supervision. Unlike prior works (\eg, TRUMANS, ZeroHSI), our method achieves all four capabilities simultaneously: RGB rendering, articulated object modeling, physical constraint modeling, and zero-shot generalization, notably without using 3D supervision.}
    \label{fig:teaser}
\end{figure*}

\begin{table*}[t]
  \setlength{\tabcolsep}{11pt}
    \centering
    \caption{Comparison of capabilities across different approaches. \textbf{HOI}: Human-object interaction; \textbf{RGB}: RGB rendering; \textbf{Art}: Articulated objects; \textbf{Phy}: Physical constraints; \textbf{ZS}: Zero-shot.}
    \label{tab:method-comparison}
    \begin{tabular}{lccccc|lccccc}
    \toprule
    \textbf{Method} & \textbf{HOI} & \textbf{RGB} & \textbf{Art} & \textbf{Phy} & \textbf{ZS} & \textbf{Method} & \textbf{HOI} & \textbf{RGB} & \textbf{Art} & \textbf{Phy} & \textbf{ZS}\\
    \midrule
    {CHOIS}~\cite{CHOIS} & $\textcolor{green}{\usym{2713}}$ & $\textcolor{red}{\usym{2717}}$ & $\textcolor{red}{\usym{2717}}$ & $\textcolor{green}{\usym{2713}}$ & $\textcolor{red}{\usym{2717}}$ &
    {Nifty}~\cite{Nifty} & $\textcolor{green}{\usym{2713}}$ & $\textcolor{red}{\usym{2717}}$ & $\textcolor{red}{\usym{2717}}$ & $\textcolor{green}{\usym{2713}}$ & $\textcolor{red}{\usym{2717}}$ \\
    {LINGO}~\cite{LINGO} & $\textcolor{green}{\usym{2713}}$ & $\textcolor{red}{\usym{2717}}$ & $\textcolor{red}{\usym{2717}}$ & $\textcolor{green}{\usym{2713}}$ & $\textcolor{red}{\usym{2717}}$  &
    {TRUMANS}~\cite{TRUMANS} & $\textcolor{green}{\usym{2713}}$ & $\textcolor{red}{\usym{2717}}$ & $\textcolor{red}{\usym{2717}}$ & $\textcolor{green}{\usym{2713}}$ & $\textcolor{red}{\usym{2717}}$ \\
    {GenZi}~\cite{GenZi} & $\textcolor{green}{\usym{2713}}$ & $\textcolor{red}{\usym{2717}}$ & $\textcolor{red}{\usym{2717}}$ & $\textcolor{red}{\usym{2717}}$ & $\textcolor{green}{\usym{2713}}$  &
    {ZeroHSI}~\cite{ZeroHSI} & $\textcolor{green}{\usym{2713}}$ & $\textcolor{red}{\usym{2717}}$ & $\textcolor{red}{\usym{2717}}$ & $\textcolor{red}{\usym{2717}}$ & $\textcolor{green}{\usym{2713}}$ \\
    {InterDreamer}~\cite{InterDreamer} & $\textcolor{green}{\usym{2713}}$ & $\textcolor{red}{\usym{2717}}$ & $\textcolor{red}{\usym{2717}}$ & $\textcolor{red}{\usym{2717}}$ & $\textcolor{green}{\usym{2713}}$  & 
    {Chao et al.}~\cite{chao2025part} & $\textcolor{red}{\usym{2717}}$ & $\textcolor{red}{\usym{2717}}$ & $\textcolor{green}{\usym{2713}}$ & $\textcolor{green}{\usym{2713}}$ & $\textcolor{red}{\usym{2717}}$ \\
    {Song et al.}~\cite{song2024reacto} & $\textcolor{red}{\usym{2717}}$ & $\textcolor{red}{\usym{2717}}$ & $\textcolor{green}{\usym{2713}}$ & $\textcolor{green}{\usym{2713}}$ & $\textcolor{red}{\usym{2717}}$  &
    \textbf{Ours} & \textbf{$\textcolor{green}{\usym{2713}}$} & \textbf{$\textcolor{green}{\usym{2713}}$} & \textbf{$\textcolor{green}{\usym{2713}}$} & \textcolor{green}{\usym{2713}} & \textcolor{green}{\usym{2713}} \\
    \bottomrule
    \end{tabular}
\end{table*}
\IEEEPARstart{S}{ynthesizing} realistic human motions that interact with 3D environments is fundamental to computer graphics, VR/AR, embodied AI, and robotics applications~\cite{CHOIS,Nifty,LINGO,TRUMANS,gao2020interactgan,diller2024cg,li2024task,xu2023interdiff,fan20253d,wang2026multimodal}. 
A substantial amount of research has been devoted to human-object interaction (HOI) synthesis over the years. While significant progress has been made in synthesizing human motions with rigid objects, interactions involving articulated objects (\eg, opening doors or cabinets) remain a challenging and under-explored problem.
The complexity arises from the part-wise kinematic constraints and motion dependencies inherent in articulated structures, which are difficult to capture under monocular settings without 3D supervision.

Recent zero-shot approaches leverage pretrained video diffusion models as motion priors to generate 4D human-object interactions without 3D/4D ground truth~\cite{ZeroHSI,InterDreamer,GenZi,liu2026open}.
However, these methods are inherently limited to rigid object manipulation, treating dynamic objects as single rigid bodies and failing to model complex part-wise articulation.
More critically, they generate interactions end-to-end from 2D priors without explicit 4D geometric reconstruction, leading to physically implausible or geometrically inconsistent results.
While recent advances in 4D scene reconstruction~\cite{Stereo4D,Feature4X,DBLP:conf/cvpr/YaoZW25,DBLP:journals/corr/abs-2504-07961,DBLP:journals/corr/abs-2507-23782,DBLP:conf/cvpr/WuLHQZD25,yuan2025self,DBLP:journals/corr/abs-2504-13152} and video-to-4D generation~\cite{DBLP:conf/cvpr/0001ZNMSVSTW025,chu2025robust,DBLP:journals/corr/abs-2507-23785,DBLP:journals/corr/abs-2502-08377,DBLP:conf/eccv/WuYJCWB24,DBLP:conf/eccv/BahmaniLYSRLLPTWTL24,wu2025cat4d,DBLP:journals/corr/abs-2503-16396} have shown promise in temporally coherent 4D representations, they primarily focus on rigid scenes or objects dynamics, leaving articulated human-object interactions largely unexplored. 
As summarized in Table~\ref{tab:method-comparison}, no existing zero-shot method can synthesize human interactions with articulated objects while maintaining physical constraints and temporal coherence.

To overcome these challenges, we propose \methodname, a zero-shot framework that synthesizes articulated human-object interactions by reconstructing 4D scene dynamics from monocular video priors.
Unlike end-to-end generation methods, we formulate the synthesis task as a 4D reconstruction problem: we first generate a 2D video from a text prompt using a diffusion model, then reconstruct a temporally coherent and geometrically consistent 4D scene through inverse rendering.
This reconstruction-based approach allows us to explicitly model part-wise articulation and human-object contact, effectively resolving the monocular ambiguity that plagues joint optimization methods.

Our framework employs a decoupled two-stage pipeline:
In the first stage, we reconstruct articulated object dynamics using optical flow-based part segmentation and kinematic constraints, leveraging motion continuity as a reliable geometric cue.
In the second stage, we synthesize human motion conditioned on the reconstructed object states, using the recovered articulation and contact geometry as physical scaffolds to ensure plausibility.
This separation enables stable optimization and enhances both geometric consistency and physical realism under monocular constraints.
Extensive experiments demonstrate that \methodname significantly outperforms existing methods in articulation accuracy, contact consistency, and physical plausibility, thereby enabling zero-shot synthesis of articulated interactions beyond the capability of prior rigid-object approaches. In summary, our key contributions include:

\begin{itemize}
\item We present the first zero-shot framework for articulated human-object interaction synthesis via 4D reconstruction, extending video-prior-driven HOI beyond rigid manipulation.
\item We propose a two-stage reconstruction-informed pipeline that recovers articulated human-object dynamics from monocular 2D video priors without 3D supervision, addressing three key challenges: articulation modeling from monocular video, monocular input ambiguity, and physical-aware synthesis.
\item We comprehensively demonstrate that \methodname enhances physical plausibility and stability, yielding lower penetration and more coherent articulated motions across diverse interaction scenarios.
\end{itemize}

%% file: sections/2_related_work.tex
\section{Related Work}
\label{sec:related_work}

\subsection{Human-Object Interaction Synthesis}

Synthesizing plausible human motions while manipulating objects has long been studied in computer animation, robotics, and embodied AI~\cite{gao2020interactgan,diller2024cg,xu2023interdiff,fan20253d,li2023object,behave,chen2026v}.
A substantial amount of research on this topic has been conducted over the years. In the early years, many methods relied on motion capture (mocap) datasets paired with object trajectories to enable data-driven synthesis of interactions~\cite{Nifty,TRUMANS,DBLP:conf/cvpr/LiuZXTYY25,lu2025humoto}. These approaches can produce physically plausible results when sufficient paired 3D scene and mocap data are available, but they require expensive capture setups and struggle to generalize to novel objects or interaction types.

In recent years, learning-based methods have emerged as the mainstream. Methods such as CHOIS~\cite{CHOIS}, LINGO~\cite{LINGO}, and InteractAnything~\cite{zhang2025interactanything} generate interactions from language prompts and sparse waypoints, yet they require training on interaction-specific data and assume known object kinematics~\cite{huang2025hunyuanvideo,li2025genhoi,jiang2023full}. These models exhibit limited generalization due to their dependence on curated motion sequences and scene-object configurations. Additionally, work on future interaction prediction~\cite{DBLP:conf/cvpr/AshutoshPG25} and hand-object motion recovery~\cite{yu2025dyn} has explored temporal dynamics in interactions, but often under constrained settings.

In contrast, zero-shot methods circumvent the data dependency by leveraging external priors from pretrained foundation models. GenZi~\cite{GenZi} generates static human poses using 2D diffusion models conditioned on scene layout. ZeroHSI~\cite{ZeroHSI} synthesizes dynamic 3D human-object interactions by distilling image-to-video model outputs into 4D Gaussian representations. However, existing zero-shot approaches assume only 6D rigid object manipulation, treating dynamic objects as single rigid bodies and failing to model part-wise articulation. This limitation prevents them from handling common interactions such as opening doors, drawers, or cabinets, where the articulated structure of objects plays a central role.

\subsection{Articulated Object Reconstruction}

Reconstructing articulated object structure and motion from visual inputs is a fundamental problem in 3D vision~\cite{chao2025part,song2024reacto,zhai2025taga,yao2025riggs,guo2025articulatedgs,lin2025splart,foundationpose,chen2026motion,liang2026dextercap,gupta2026pokenet,kim2026camo}.
These methods can be broadly classified into two categories. The first category relies on category-level templates or known part hierarchies. Methods like Reacto~\cite{song2024reacto}, TAga~\cite{zhai2025taga}, and RIGGS~\cite{yao2025riggs} recover articulated object kinematics from monocular videos by leveraging predefined part decompositions or kinematic chains. While effective for known object categories, they limit applicability to novel objects and require multi-view inputs in many cases. D3D-HOI~\cite{xu2021d3dhoi} and 3DADN~\cite{qian2022understanding} extend articulated reconstruction to human-object interaction settings but still operate under object-centric assumptions.
The second category adopts unsupervised paradigms to discover articulated parts from motion cues alone~\cite{deng2024articulate,peng2025generalizable,xu2021articulated,goyal2025geopard,zhang2025adaptive}. By analyzing optical flow, scene flow, or point cloud motion, these methods identify rigid parts and infer joint axes without category-specific templates. However, they operate purely on object-centric motion and ignore the rich geometric and physical signals provided by human-object interaction. When a human hand grasps a door handle and pushes, the contact region and motion trajectory provide strong priors for articulation inference that are largely unexploited. In contrast, we propose to leverage human interaction as a substantial prior for part discovery by exploiting contact cues and temporal motion coherence, enabling structure inference without category-specific templates.

\subsection{Video Distillation for 3D Reconstruction}

Recent zero-shot 3D methods leverage video diffusion models (VDMs) as powerful priors to generate 4D human-scene interactions without 3D supervision.
A substantial amount of work has been devoted to distilling VDM outputs into consistent 4D representations. Methods like Zero4D~\cite{park2025zero4d} and Free4D~\cite{liu2025free4d} show that a single input video can be extended into coherent 4D sequences by sampling from VDMs, while VideoScene~\cite{wang2025videoscene} distills these outputs directly into 3D Gaussians in a single forward pass. Building on this foundation, numerous approaches have advanced 4D reconstruction from monocular videos~\cite{Stereo4D,Feature4X,DBLP:conf/cvpr/YaoZW25,DBLP:journals/corr/abs-2504-07961,DBLP:journals/corr/abs-2507-23782,DBLP:conf/cvpr/WuLHQZD25,yuan2025self,DBLP:journals/corr/abs-2504-13152,4DLangSplat,DBLP:conf/eccv/MihajlovicPTMBTB24,DBLP:conf/cvpr/MatsukiBD25}, video-to-4D generation~\cite{DBLP:conf/cvpr/0001ZNMSVSTW025,chu2025robust,DBLP:journals/corr/abs-2507-23785,DBLP:journals/corr/abs-2502-08377,DBLP:conf/eccv/WuYJCWB24,DBLP:conf/eccv/BahmaniLYSRLLPTWTL24,wu2025cat4d,DBLP:journals/corr/abs-2503-16396}, and specialized applications~\cite{DBLP:conf/cvpr/TaubnerZTL25,DBLP:conf/cvpr/ZhaoNWZZW0CWZMW25,DBLP:conf/cvpr/PengZWXXZKTZ25,DBLP:conf/cvpr/KwonC025,DNF,DIO,Mamba4D,DBLP:conf/cvpr/SangCCMBC25,DBLP:journals/corr/abs-2503-09631,DBLP:conf/cvpr/LiuYL0Z25}. However, these methods primarily focus on rigid scenes or simple object motions; their performance degrades when the scene contains articulated objects with part-wise kinematics.

Recent work has also explored diffusion-based generation of articulated objects~\cite{li2025articulated,zhang2025physrig,kreber2025guiding,su2025artformer,gao2025meshart}. Li et al.~\cite{li2025articulated} address articulated object kinematics by distilling motion patterns from video diffusion models. Nevertheless, despite their success in modeling rigid-body dynamics or biological articulation, these approaches often treat objects as monolithic entities, applying a single global transformation across the entire object, or require additional supervision for part decomposition. They fail to address the core challenge of jointly synthesizing articulated human-object interactions from monocular video priors, where both human motion and object articulation must be recovered in a geometrically consistent and physically plausible manner under severe monocular ambiguity.

%% file: sections/3_method.tex
\section{Methodology}
\label{sec:method}

\subsection{Problem Formulation and Overview}
\label{subsec:overview}
We address the problem of synthesizing physically plausible articulated human-object interactions from monocular video priors \textbf{without any 3D supervision}.
Existing zero-shot methods~\cite{ZeroHSI} treat all objects as rigid entities and thus fail to model part-wise articulation (\eg, doors, drawers, cabinets).
A natural direction is to directly generate 3D interactions through end-to-end differentiable rendering. However, this approach suffers from a fundamental ambiguity. Under monocular observation, it is unclear whether motion in the image arises from human movement, object articulation, or their combination.
Jointly optimizing human and object dynamics in a single stage leads to conflicting gradients and unstable convergence, as the optimization landscape couples two inherently different motion modalities with weak 2D supervision.

We instead formulate interaction synthesis as a \textbf{4D reconstruction problem}. Given a monocular video $\mathcal{V} = \{I(t)\}_{t=1}^T$ (generated from a text prompt $\mathcal{T}$ using video diffusion models or captured from real scenes), we synthesize 3D interactions by reconstructing a full 4D articulated scene through inverse rendering, using the 2D video as supervision.
This reconstruction-based view injects explicit geometric and kinematic priors into the monocular dynamics, allowing us to disambiguate human and object motion through structured constraints rather than relying on data-driven heuristics.
The human is parameterized by SMPL-X~\cite{SMPLX, SMPL} with shape $\boldsymbol{\beta} \in \mathbb{R}^{10}$, pose $\boldsymbol{\psi}(t) \in \mathbb{R}^{J \times 3}$, and translation $\boldsymbol{\tau}(t) \in \mathbb{R}^3$. Articulated object parts are governed by rigid $SE(3)$ transformations $\mathbf{T}^d(t)$.
We represent both human and object using 3D Gaussians~\cite{3dgs} for end-to-end optimization.

\begin{figure*}[t]
    \centering
    \includegraphics[width=0.99\linewidth]{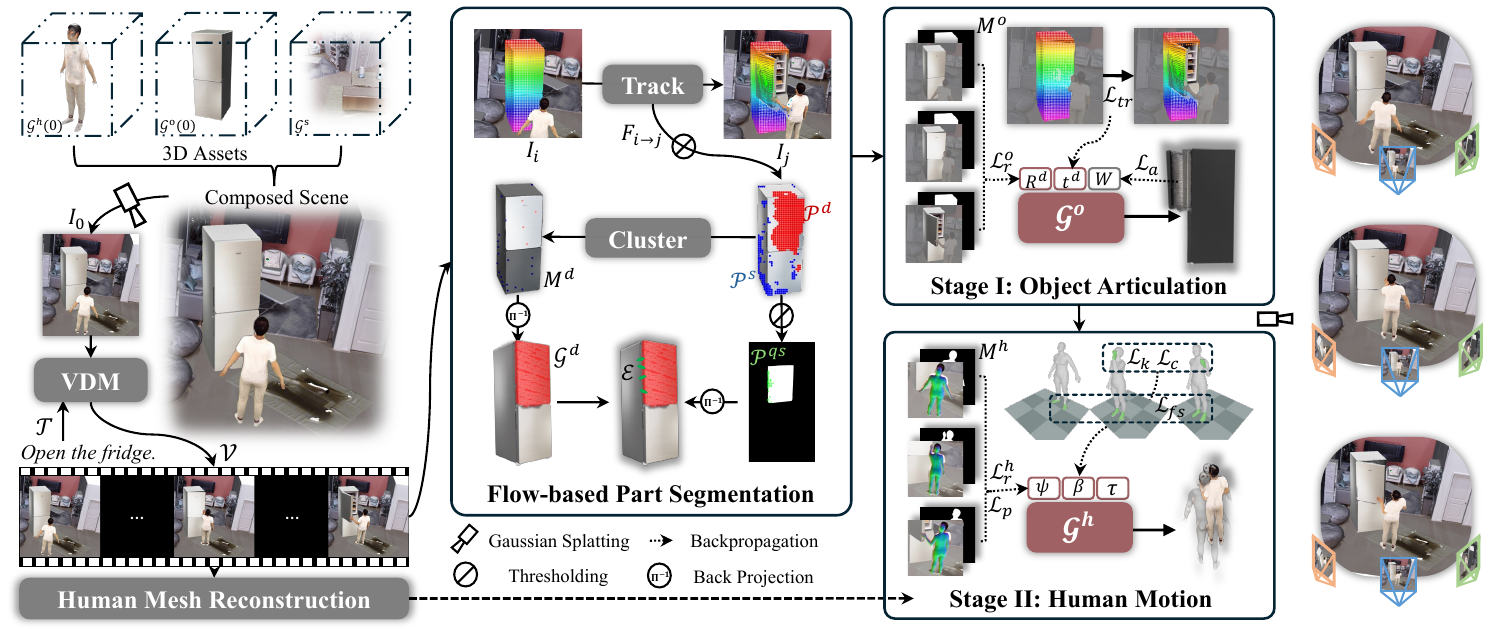}
    \caption{
    \methodname synthesizes 3D articulated interactions by reconstructing 4D scenes from monocular video priors. Stage I reconstructs object articulation with kinematic constraints. Stage II refines human motion under the reconstructed geometry. 
    }
    \label{fig:pipeline}
\end{figure*}

As illustrated in Fig.~\ref{fig:pipeline}, we employ a \textbf{decoupled two-stage reconstruction framework}.  
(1) In \textit{Stage I} (Sec.~\ref{subsec:flow-segmentation} to \ref{subsec:stage1}), we first identify articulated object parts via flow-based segmentation, then recover their 3D articulation through optimization with kinematic constraints, establishing a geometrically consistent 4D object scaffold.  
(2) In \textit{Stage II} (Sec.~\ref{subsec:stage2}), we refine human motion conditioned on this scaffold, using the recovered object articulation and contact geometry as hard constraints to ensure physically plausible interactions.
By solving object articulation before human motion, we avoid the ambiguity of joint optimization. The object stage has clear kinematic structure (rigid parts, hinge-like motion), while the human stage receives a fixed, physically coherent reference from which to derive contact targets.

\subsection{Flow-based Part Segmentation}
\label{subsec:flow-segmentation}

Identifying which object regions are articulated (\eg, door panel vs.\ frame) is essential for kinematic modeling but challenging under monocular observation. Neither appearance nor a single frame provides reliable cues for part boundaries.
We argue that \textit{motion} is the most reliable signal. Static regions (\eg, cabinet frame) exhibit near-zero 2D displacement across frames, while articulated regions (\eg, door panel) move with the human.
We present a flow-based segmentation pipeline that combines point tracking, SAM-guided masks, back projection to 3D, and quasi-static binding (\Cref{fig:pipeline} middle).

\noindent \textbf{Point Tracking and Motion Classification.}
We use a pre-trained point tracking network~\cite{co-tracker} that produces dense 2D trajectories across frames.
Given the video and per-frame object masks $M^o(t)$, we sample points on the object and track them from a source frame to a target frame. The displacement $\Delta p = p_{\text{tgt}} - p_{\text{src}}$ gives the flow magnitude at each point.
Points with $\|\Delta p\|_2 > \tau_f$ are classified as dynamic (articulated part), and points with $\|\Delta p\|_2 \leq \tau_f$ as static.
We sample source and target frames with sufficient temporal separation so that articulated motion is measurable while avoiding drift.
To obtain prompts for SAM, we cluster the dynamic and static point sets (\eg, via $k$-means) and use cluster centers as prompts, reducing noise and ensuring spatial coverage.

\noindent \textbf{SAM-guided Dense Mask.}
Optical flow yields sparse point-level labels. We need dense, boundary-accurate masks for 3D assignment.
We feed the clustered dynamic and static points as positive and negative prompts to Segment Anything (SAM)~\cite{SAM, SAM2}, which produces a binary mask $M^d(t)$ separating articulated from static object parts, given by
\begin{equation}
M^d(t) = \text{SAM}(I(t), \mathcal{P}^d, \mathcal{P}^s).
\end{equation}
We render the object from the canonical pose to obtain a clean image for SAM, avoiding occlusions from the human.

\noindent \textbf{Back Projection to 3D Gaussians.}
The 2D SAM mask must be transferred to the 3D Gaussian representation.
For each pixel $p$ in $M^d(t)$, we find its $K$ nearest object Gaussians in 2D image space.
We compute a soft influence of each pixel on each Gaussian using a splatting-style accumulation, where the influence depends on the 2D distance $\|p - \Pi(\mathbf{g})\|$, the Gaussian's projected opacity $\alpha$, and depth ordering (closer Gaussians receive more weight than occluded ones).
Aggregating influences over all pixels in the dynamic mask yields a per-Gaussian score $s_i^d$, and aggregating over the static mask yields $s_i^s$.
Gaussians with $s_i^d > s_i^s$ (or above a threshold) are assigned to $\mathcal{G}^d$, others to $\mathcal{G}^s$.
To resolve ambiguous or isolated assignments (\eg, Gaussians at part boundaries or small clusters), we apply a connectivity refinement. We build a $k$-nearest-neighbor graph on the 3D Gaussian positions, compute the largest connected component within each of the dynamic and static sets, and reassign remaining Gaussians to the closer component centroid in 3D.
This produces clean, spatially coherent dynamic or static partitions.

\noindent \textbf{Quasi-static Binding.}
To enforce rigid-body constraints, we must link dynamic and static parts at the articulation boundary.
Articulation points (\eg, door hinges) lie at the interface. They belong to the dynamic region but exhibit relatively low motion because they rotate rather than translate.
Within the dynamic 2D points, we identify quasi-static points as those with motion magnitude below a percentile of the overall motion distribution.
We then find their corresponding 3D Gaussians. Quasi-static points project onto a subset of dynamic Gaussians near the boundary.
For each such quasi-static dynamic Gaussian, we find its nearest static Gaussian within a 3D radius $r$. Each such pair forms a binding constraint.
\begin{equation}
    \begin{aligned}
        \mathcal{E} &= \big\{[\mathbf{g}^{qs}, \mathbf{g}^{st}] \mid \mathbf{g}^{qs} \in \mathcal{G}^d,\, \Pi(\mathbf{g}^{qs}) \in \mathcal{P}^{qs}, \\
        &\qquad \mathbf{g}^{st} \in \mathcal{G}^s,\, \|\mathbf{g}^{qs} - \mathbf{g}^{st}\|_2 \leq r\big\},
    \end{aligned}
\end{equation}
where $\mathbf{g}^{qs}$ is a quasi-static dynamic Gaussian and $\mathbf{g}^{st}$ is its bound static neighbor.

\subsection{Decoupled Two-Stage Reconstruction}
\label{subsec:two-stage-framework}
We employ a two-stage reconstruction strategy that decouples object articulation from human motion refinement.
In joint optimization, gradients from reconstruction loss compete with those from kinematic and contact terms. The human and object parameters are coupled through the shared 2D supervision, leading to unstable convergence and geometric inconsistencies.
By solving object articulation first, we obtain a fixed 4D scaffold. Stage II then refines human motion under this scaffold without modifying object geometry.
The coupling between human actions and object articulations is preserved implicitly. Stage I is supervised by the full video (including human motion), and Stage II uses the reconstructed object to derive contact targets.


\noindent \textbf{Stage I: Object Articulation Reconstruction.}
\begin{figure*}[t]
    \centering
    \includegraphics[width=0.99\linewidth]{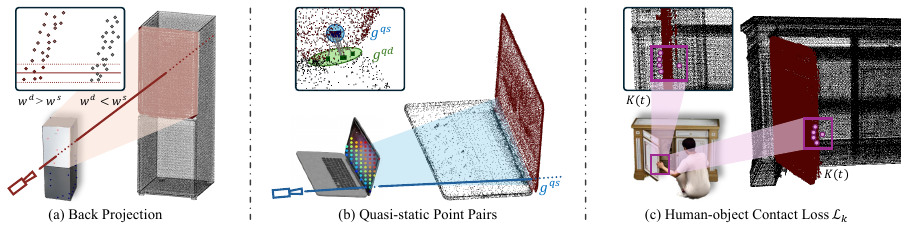}
    \caption{
    Key components for articulated interaction under monocular supervision. (a) Back projection maps masks to 3D to identify moving parts. (b) Quasi-static point pairs link dynamic/static regions for kinematic stability. (c) Contact loss projects 2D keypoints into 3D using object depth, guiding human motion without multi-view cues. Ablations in~\Cref{fig:ablation} (middle: (b), right: (c)).}
    \label{fig:projection}
\end{figure*}
\label{subsec:stage1}
We reconstruct articulated object motion by optimizing $SE(3)$ transformations $\mathbf{T}^d(t)$ per frame, using the flow-based segmentation to assign each Gaussian to dynamic or static parts (\Cref{fig:pipeline} right top).
Object articulation is inherently better constrained than human-object joint optimization. The kinematic structure (rigid parts, hinge-like motion) provides strong regularization, and optical flow supplies direct 2D motion supervision for the dynamic region.

We represent articulated motion as $\mathbf{T}^d(t) = [\mathbf{R}^d(t), \mathbf{t}^d(t)]$ with $\mathbf{R}^d(t) \in SO(3)$ and $\mathbf{t}^d(t) \in \mathbb{R}^3$.
We also introduce articulation weights $\mathbf{W}^o \in \mathbb{R}^{V \times J}$ for each part, determined from the flow-based segmentation and fixed during optimization.
The object Gaussians are driven by articulation parameters,
\begin{equation}
\boldsymbol{\mu}^o_i(t) = w_{i}^d \mathbf{T}^d(t) \boldsymbol{\mu}_i^o(0) + w_{i}^s \boldsymbol{\mu}_i^o(0)
\end{equation}
where $w_{i}^d$ and $w_{i}^s$ are object articulation weights for the articulated part and static components, respectively.
The optimization objective integrates four complementary losses,
\begin{equation}
\min_{\{\mathbf{R}^d, \mathbf{t}^d\}} \mathcal{L}_r^o + \lambda_a \mathcal{L}_a + \lambda_s \mathcal{L}_s + \lambda_{tr} \mathcal{L}_{tr}.
\end{equation}
\textbf{Reconstruction loss} $\mathcal{L}_r^o$ drives the rendered object to match the video prior, ensuring visual fidelity:
\begin{equation}
\mathcal{L}_r^o = \|\mathcal{R}(\mathcal{G}^o(t)) - I(t)\|_2^2 + \beta^o \|\mathcal{S}(\mathcal{G}^o(t)) - M^o(t)\|_2^2,
\end{equation}
where $\mathcal{R}(\cdot)$ denotes differentiable rendering and $\mathcal{S}(\cdot)$ extracts silhouette via alpha thresholding.
Without kinematic regularization, reconstruction alone can produce implausible articulations (e.g., parts drifting apart).

\begin{equation}
\begin{aligned}
\mathcal{L}_a &= \sum_{(\mathbf{g}^d, \mathbf{g}^s) \in \mathcal{E}} \|d(\mathbf{g}^d(t), \mathbf{g}^s(t)) - d(\mathbf{g}^d(0), \mathbf{g}^s(0))\|_2^2, \\
\mathcal{L}_{tr} &= \sum_{i \in \mathcal{P}_{\text{dyn}}} \left\| \hat{p}_{\text{tgt}}^i - p_{\text{tgt}}^i \right\|_2^2,
\end{aligned}
\end{equation}
where $d(\cdot, \cdot)$ is Euclidean distance, $(\mathbf{g}^d, \mathbf{g}^s)$ ranges over binding pairs in $\mathcal{E}$, $\mathcal{P}_{\text{dyn}}$ is the set of dynamic particles, $p_{\text{tgt}}^i$ is the tracker output, and $\hat{p}_{\text{tgt}}^i$ is the predicted 2D position from weighted projection of influencing dynamic Gaussians.
The tracking loss $\mathcal{L}_{tr}$ aligns the 2D projection of dynamic Gaussians with point-tracker trajectories.
We randomly sample a source-target frame pair $(t_{\text{src}}, t_{\text{tgt}})$ with sufficient temporal separation so that articulated motion is measurable.
On the object mask at $t_{\text{src}}$, we sample a set of particles $\mathcal{P}$ and track them to $t_{\text{tgt}}$ using the point tracker, obtaining 2D trajectories $\{p_{\text{src}}^i \mapsto p_{\text{tgt}}^i\}$.
We classify each particle as dynamic or static based on flow magnitude $\|p_{\text{tgt}}^i - p_{\text{src}}^i\|_2$. Only particles in the moving region are used for $\mathcal{L}_{tr}$.
For each dynamic particle $p_{\text{src}}^i$, we compute its soft influence on nearby object Gaussians (via $K$-nearest neighbors in image space, weighted by projected opacity and 2D distance).
The predicted 2D position at $t_{\text{tgt}}$ is a weighted combination of the projected 3D positions of these Gaussians (transformed by $\mathbf{T}^d(t_{\text{tgt}})$), using the same influence weights.
We minimize the distance between this predicted position and the tracker output $p_{\text{tgt}}^i$.
Only Gaussians that influence particles in the dynamic region contribute to the gradient. Static Gaussians are excluded since they have no expected motion.
\textbf{Smoothness loss} $\mathcal{L}_s$ penalizes abrupt changes in $\mathbf{T}^d(t)$ across frames, encouraging temporally coherent articulation trajectories.

We optimize Stage I frame-by-frame.
For the first frame, we initialize $\mathbf{T}^d(0)$ as the identity (no articulation).
For subsequent frames, we warm-start from the previous frame. $\mathbf{T}^d(t)$ is initialized from $\mathbf{T}^d(t-1)$, or from linear extrapolation when $t > 1$, so that the optimization continues from a temporally consistent solution.

\begin{algorithm}[t]
\caption{Stage I: Object Articulation Reconstruction}
\label{alg:stage1}
\begin{algorithmic}
\STATE \textbf{Input:} Video $\mathcal{V}$, object masks $M^o(t)$, flow-based segmentation $\mathcal{G}^d$, $\mathcal{G}^s$, $\mathcal{E}$
\STATE \textbf{Output:} Articulated transforms $\mathbf{T}^d(t)$ for $t = 1, \ldots, T$
\STATE Initialize $\mathbf{T}^d(0) \gets \mathbf{I}$
\FOR{$t = 1$ to $T$}
\STATE Warm-start $\mathbf{T}^d(t)$ from $\mathbf{T}^d(t-1)$
\STATE Sample source-target pair $(t_{\text{src}}, t_{\text{tgt}})$; track particles $\mathcal{P}$ via point tracker
\STATE Classify dynamic/static particles by flow magnitude; obtain $\mathcal{P}_{\text{dyn}}$
\FOR{each iteration}
\STATE Compute $\mathcal{L}_r^o$, $\mathcal{L}_a$, $\mathcal{L}_{tr}$, $\mathcal{L}_s$
\STATE Update $\mathbf{T}^d(t)$ via Adam
\ENDFOR
\ENDFOR
\STATE \textbf{return} $\{\mathbf{T}^d(t)\}_{t=1}^T$
\end{algorithmic}
\end{algorithm}

\noindent \textbf{Stage II: Human Motion Refinement.}
\label{subsec:stage2}
With the object articulation fixed from Stage I, we refine human motion under the reconstructed 4D geometry (\Cref{fig:pipeline} right bottom).
The key challenge is obtaining 3D contact targets. Without multi-view input, we cannot directly observe where the human hands contact the object in 3D.
\textbf{3D Contact Keypoint Derivation.}
We derive $\mathcal{K}_t$ and $\mathbf{K}_j(t)$ from 2D evidence as follows (see~\Cref{fig:projection} (c)).
\textbf{(1) Contact frame selection.} We identify frames where the object is articulating by monitoring changes in $\mathbf{T}^d(t)$ between consecutive frames.
Significant rotation or translation indicates active human-object contact.
We apply temporal smoothing (\eg, max-pooling over a short window) to suppress spurious detections from minor jitter.
\textbf{(2) 2D contact region.} At each contact frame, we render the object's silhouette from the reconstructed 3D Gaussians and compare it with the human and object SAM masks.
The key observation is that pixels where the \textit{human} mask overlaps the \textit{object model} silhouette, but the \textit{object SAM} mask is absent, correspond to human body parts occluding the object from the camera's view. That is, the hand (or other limb) is in front of the object surface at that pixel.
We thus define the contact region as $M^h(t) \cap \mathcal{S}(\mathcal{G}^o(t)) \setminus M^o_{\text{sam}}(t)$, where the object model silhouette comes from Stage I and the human/object masks from segmentation.
\textbf{(3) Joint assignment.} We use 2D keypoints from the initial human pose estimate (GVHMR~\cite{GVHMR} or similar) and project SMPL-X joints to the image.
Joints whose 2D locations fall inside the contact region and whose confidence exceeds a threshold are retained.
Hand and fingertip joints (SMPL-X indices for left/right hands) dominate, as they are the primary contact points during manipulation.
\textbf{(4) 3D lifting.} For each retained 2D keypoint, we find the $K$ nearest \textit{dynamic} object Gaussians in image space (the articulated part, not the static frame).
Among these, we select the Gaussian with minimum depth (the one closest to the camera) as the object surface point the hand contacts.
We use this Gaussian's 3D position, optionally offset slightly toward the camera along the view ray to avoid penetration, as the contact target $\mathbf{K}_j(t)$.
This yields a sparse set of 3D contact keypoints per frame for the kinematic loss.
We optimize SMPL-X parameters $\boldsymbol{\theta}(t)$ to drive human Gaussians $\mathcal{G}^h(t)$, matching hand joints to $\mathcal{K}_t$ while preserving natural motion and avoiding penetration.
The optimization objective is:
\begin{equation}
\min_{\boldsymbol{\theta}} \mathcal{L}_r^h + \lambda_p \mathcal{L}_p + \lambda_{fs} \mathcal{L}_{fs} + \lambda_s \mathcal{L}_s + \lambda_k \mathcal{L}_k + \lambda_c \mathcal{L}_c.
\end{equation}
\textbf{Reconstruction loss} $\mathcal{L}_r^h$ ensures the rendered human aligns with the video,
\begin{equation}
\mathcal{L}_r^h = \sum_{t=1}^T \|\mathcal{R}(\mathcal{G}^h(t)) - I(t)\|_2^2 + \beta^h \|\mathcal{S}(\mathcal{G}^h(t)) - M^h(t)\|_2^2,
\end{equation}
\textbf{Kinematic loss} $\mathcal{L}_k$ pulls hand joints toward the 3D contact keypoints derived from Stage I, ensuring proper hand-object contact,
\begin{equation}
\mathcal{L}_k = \sum_{t=1}^T \sum_{j \in \mathcal{K}_t} \|\mathbf{J}_j(\boldsymbol{\theta}(t)) - \mathbf{K}_j(t)\|_2^2,
\end{equation}
where $\mathcal{K}_t$ indexes joints with confident contact and $\mathbf{K}_j(t)$ is the corresponding 3D target.
\textbf{Prior loss} $\mathcal{L}_p$ regularizes the refined motion toward the initial VDM-estimated pose $\boldsymbol{\theta}_v(t)$, preventing overfitting to contact targets at the expense of naturalness.
\textbf{Foot sliding loss} $\mathcal{L}_{fs}$ prevents unrealistic horizontal foot movement during ground contact.
We use GVHMR~\cite{GVHMR} to obtain per-frame foot contact estimates (left/right ankle and foot).
For each foot, we identify temporal intervals where contact is sustained. Within each interval, we penalize deviation of the foot vertices from their mean position, encouraging the foot to remain stationary while in contact.
\textbf{Collision loss} $\mathcal{L}_c$ penalizes penetration between the human mesh and the object.
We sample vertices $\mathcal{V}_h$ from the hand regions (SMPL-X hand mesh) and points $\mathcal{Q}^o$ from the object surface (dynamic Gaussians or mesh).
For each pair $(v, q)$ with $v \in \mathcal{V}_h$ and $q \in \mathcal{Q}^o$, we compute the distance $d_{vq} = \|\mathbf{v}(t) - \mathbf{q}(t)\|_2$.
A hinge loss penalizes distances below a threshold $\delta$,
\begin{equation}
\mathcal{L}_c = \sum_{t=1}^T \sum_{v \in \mathcal{V}_h} \sum_{q \in \mathcal{Q}^o} \max(0, \delta - d_{vq}),
\end{equation}
so that the optimizer pushes hand vertices away from the object when they are too close.
The prior and foot sliding terms are:
\begin{equation}
\begin{aligned}
\mathcal{L}_p &= \sum_{t=1}^T \|\boldsymbol{\theta}(t) - \boldsymbol{\theta}_v(t)\|_2^2 + \eta \|\boldsymbol{\psi}(t) - \boldsymbol{\psi}_v(t)\|_2^2, \\
\mathcal{L}_{fs} &= \sum_{\text{contact intervals}} \sum_{t \in I} \sum_{v \in \mathcal{V}_{foot}} \|\mathbf{v}(t) - \bar{\mathbf{v}}_I\|_2^2,
\end{aligned}
\end{equation}
where $\boldsymbol{\theta}_v(t)$ and $\boldsymbol{\psi}_v(t)$ are the VDM-estimated parameters, $\mathcal{V}_{foot}$ denotes foot vertices, and $\bar{\mathbf{v}}_I$ is the mean foot position over contact interval $I$.

\begin{algorithm}[t]
\caption{Stage II: Human Motion Refinement}
\label{alg:stage2}
\begin{algorithmic}
\STATE \textbf{Input:} Fixed $\{\mathbf{T}^d(t)\}$, $\mathcal{G}^o(t)$, video $\mathcal{V}$, human masks $M^h(t)$
\STATE \textbf{Output:} SMPL-X parameters $\boldsymbol{\theta}(t)$ for $t = 1, \ldots, T$
\STATE Derive 3D contact keypoints $\mathcal{K}_t$, $\mathbf{K}_j(t)$ from contact region and object depth
\STATE Initialize $\boldsymbol{\theta}(t)$ from VDM-estimated pose $\boldsymbol{\theta}_v(t)$
\FOR{each iteration}
\STATE Compute $\mathcal{L}_r^h$, $\mathcal{L}_k$, $\mathcal{L}_p$, $\mathcal{L}_{fs}$, $\mathcal{L}_s$, $\mathcal{L}_c$
\STATE Update $\boldsymbol{\theta}(t)$ via Adam (jointly over all $t$)
\ENDFOR
\STATE \textbf{return} $\boldsymbol{\theta}(t)$, yielding $\mathcal{G}^h(t)$ and final scene $\mathcal{G}(t) = \mathcal{G}^h(t) \cup \mathcal{G}^o(t) \cup \mathcal{G}^s$
\end{algorithmic}
\end{algorithm}

\noindent \textbf{Temporal strategy.} For Stage I, we optimize $\mathbf{T}^d(t)$ frame-by-frame, initializing each frame from the previous frame's solution to ensure temporal continuity. The first frame uses identity transformation. For Stage II, we optimize all frames jointly. Object Gaussians $\mathcal{G}^o(0)$ are initialized in canonical space. The final scene combines $\mathcal{G}(t) = \mathcal{G}^h(t) \cup \mathcal{G}^o(t) \cup \mathcal{G}^s$ for rendering.

%% file: sections/4_experiments.tex
\section{Experiments}
\label{sec:experiments}

We conduct comprehensive experiments to evaluate our \methodname performance on zero-shot articulated human-object interaction synthesis.
Our evaluation covers two main aspects: reconstruction quality (geometric consistency, physical plausibility, temporal coherence) and articulated object dynamics accuracy.
We compare our approach against several state-of-the-art baselines and demonstrate significant improvements across multiple reconstruction metrics.

\begin{figure*}[!t]
    \centering
    \includegraphics[width=0.99\linewidth]{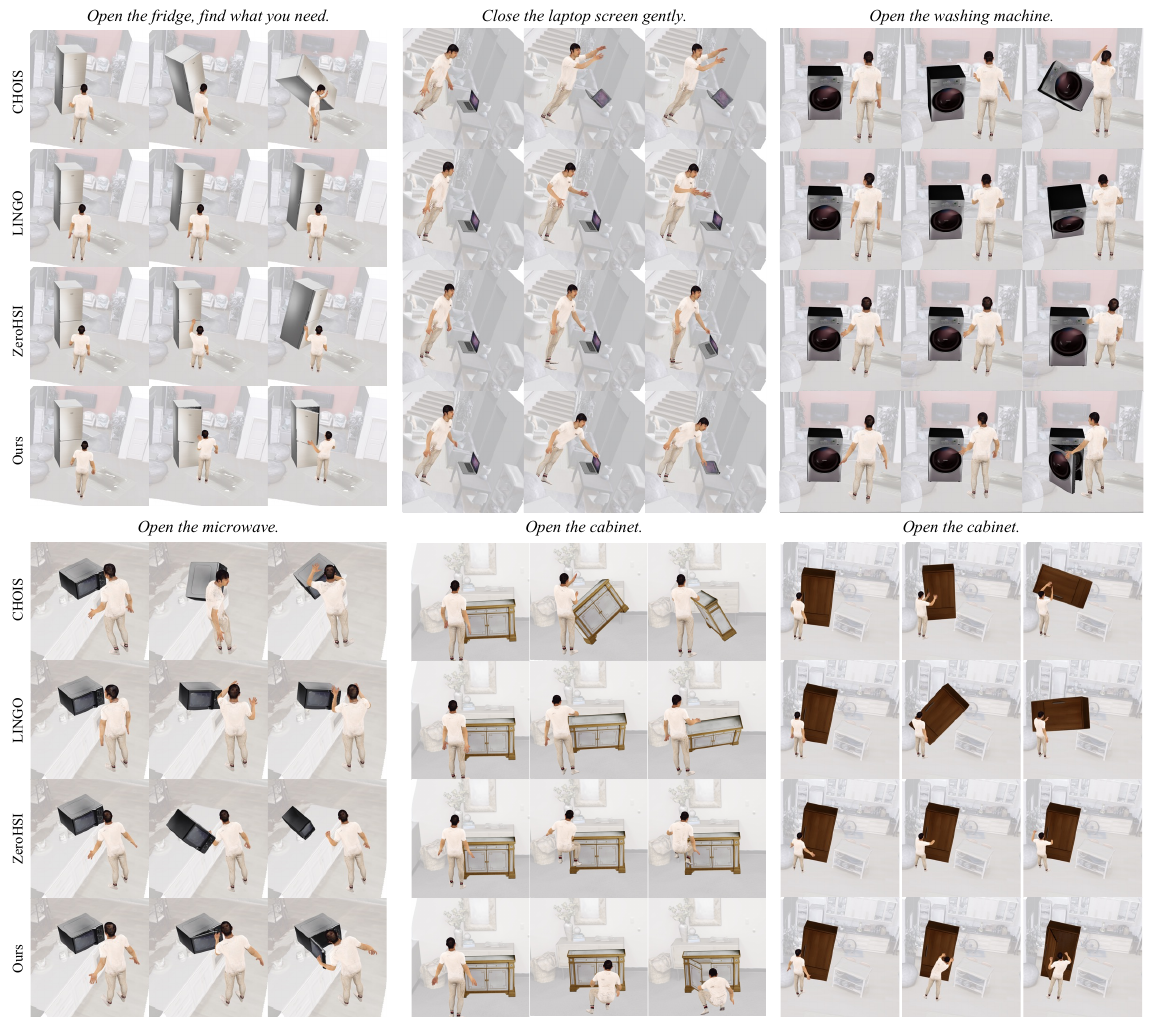}
    \caption{Qualitative comparison of our method with baselines. Our method synthesizes more realistic articulated human-object interactions with proper contact and natural motion coordination. Better inspected in our supplementary video.}
    \label{fig:qualitative_results}
\end{figure*}

\subsection{Settings}
\noindent \textbf{Baselines.}
We compare our method against four representative approaches: TRUMANS~\cite{TRUMANS}, a mocap-based method requiring paired 3D scene and motion capture data; LINGO~\cite{LINGO}, a language-guided human motion synthesis approach; CHOIS~\cite{CHOIS}, a contact-aware human-object interaction synthesis method; and ZeroHSI~\cite{ZeroHSI}, a zero-shot method leveraging video diffusion models for rigid object interactions.
Additionally, for articulated object dynamics, we compare against D3D-HOI~\cite{xu2021d3dhoi} and 3DADN~\cite{qian2022understanding}, which are designed explicitly for monocular articulated object estimation, providing a direct comparison for articulation recovery from monocular video priors.

\noindent \textbf{Datasets and Metrics.}
\textbf{1) For articulated object dynamics,} we use single-view videos rendered from scenes in the ArtGS dataset~\cite{liu2025building} with ground truth annotations.
\textbf{2) For human-object interaction}, we follow ZeroHSI~\cite{ZeroHSI} where each scene is annotated with natural language descriptions of human-scene interactions and corresponding initial positions.
The scenes are from Replicate~\cite{straub2019replica}, with humans from XHumans~\cite{shen2023xavatar} and objects generated by Trellis~\cite{trellis}.
We employ two categories of metrics: reconstruction quality and articulated object dynamics accuracy. The metrics are defined as follows. \textbf{X-CLIP Score}~\cite{xclip} measures semantic alignment between synthesized interactions and textual descriptions using cross-modal similarity. We compute similarity scores between video frames and text prompts, processing $8$ sampled frames per video sequence with a frame sample rate of $1.0$, and report the softmax probability corresponding to the ground-truth scene description; higher scores indicate better text-to-motion correspondence. \textbf{Motion Smoothness} evaluates temporal consistency by computing velocity stability and acceleration magnitude across human joint trajectories. We calculate joint velocities (first-order derivatives) and accelerations (second-order derivatives) for all SMPL-X~\cite{SMPLX, SMPL} joints at $30$ FPS; the smoothness score is the standard deviation of joint speeds across all frames, with lower values indicating smoother motion. \textbf{Foot Sliding} detects unrealistic foot movement using a mesh-based algorithm that analyzes four foot joints (left/right ankles and toes) from SMPL-X, computing distances to the ground mesh and projecting displacements onto horizontal planes perpendicular to the ground. The sliding threshold is $0.001$ m/frame; we report the ratio of sliding frames to contact frames multiplied by average sliding distance, with lower indicating more realistic contact. \textbf{Contact\%} measures the percentage of frames where hand joints (left/right wrists) maintain proper contact with object vertices; higher contact percentages indicate more consistent interaction. \textbf{Penetration\%} quantifies physical plausibility using a mesh-based penetration detection algorithm that computes distances between human vertices and scene/object meshes, using vertex normals to determine penetration direction. \textbf{Rotation errors} measure the angular difference between estimated and ground-truth joint rotations. We report mean, standard deviation, maximum, minimum, and median rotation errors across all joints and frames in degrees.

\subsection{Implementation Details}

\noindent \textbf{Training.} We implement our framework using PyTorch~\cite{pytorch} with the Adam~\cite{adam} optimizer and 3D Gaussian splatting~\cite{3dgs} for differentiable rendering. Canonical Gaussians are initialized from the first frame. Human Gaussians $\mathcal{G}^h(0)$ are distributed across the SMPL-X~\cite{SMPLX, SMPL, smplify} mesh surface; articulated objects use canonical Gaussians $\mathcal{G}^o(0)$ per part with articulation weights from flow-based segmentation. We use batch size $1$ and gradient clipping (max norm $1.0$) for training stability.
For \textbf{Flow-based segmentation}, we use CoTracker~\cite{co-tracker} for robust point tracking with temporal consistency. Flow magnitude thresholds $\tau_f^d = 5$ pixels (dynamic) and $\tau_f^s = 2$ pixels (static) distinguish articulated from static regions. SAM~\cite{SAM, SAM2} ViT-H with default parameters produces dense masks; we render the object from canonical pose as SAM input to avoid human occlusion. Quasi-static points are identified by the 10th percentile of motion magnitudes within dynamic regions (minimum $1.0$ px) to capture articulation boundaries. The binding radius is $r = 0.05$ m in 3D. We apply $k$-nearest-neighbor connectivity refinement to reassign ambiguous Gaussians at part boundaries.
For \textbf{Stage I} (Object Articulation), we use learning rate $1.0 \times 10^{-4}$ with loss weights $\lambda_r = 1.0$, $\lambda_{tr} = 2.0$, $\lambda_a = 0.05$, $\lambda_s = 1.0$. We run $200$ iterations per frame with early stopping. For \textbf{Stage II} (Human Motion), learning rates are $1.0 \times 10^{-3}$ (pose) and $1.0 \times 10^{-4}$ (camera), with $\lambda_s = \lambda_k = 1.0 \times 10^{4}$, $\lambda_p = 1.0$, $\lambda_{fs} = 10$, $\lambda_c = 1.0 \times 10^{5}$. Stage II runs $1000$ iterations.
The video diffusion model is KLing (default), with batch size $1$ and gradient clipping (max norm $1.0$). Training takes approximately $30$ minutes on an NVIDIA A6000 (48GB). For HOI synthesis, we follow ZeroHSI, with the same text prompts, scene configurations, and output lengths (60--120 frames).

\noindent \textbf{Runtime.} Our total runtime is composed of four main steps: video generation with KLing ($5$ min), flow-based segmentation ($2$ min), Stage I object articulation optimization ($15$ min), and Stage II human motion synthesis ($8$ min), yielding a total of approximately $30$ minutes on a single NVIDIA A6000 (48GB) GPU. This efficiency is achieved through our two-stage strategy, where Stage I focuses on object articulation while Stage II synthesizes human motion, allowing for parallel processing opportunities in future implementations.

\begin{table*}[!t]
  \centering
  
  \setlength{\tabcolsep}{16pt}
  \caption{
    Comparison of interaction quality. Smoothness (↓) is best interpreted among zero-shot methods. Non-zero-shot high smoothness stems from minimal contact, not motion instability.
  }
  \begin{tabular}{l|c|ccccc}
    \toprule
    \textbf{Method} & Zero-shot & X-CLIP ↑ & Smoothness ↓ & Foot Sliding ↓ & Contact\% ↑ & Penetration\% ↓ \\
    \midrule
    {TRUMANS}~\cite{TRUMANS} & $\textcolor{red}{\usym{2717}}$ & 0.169 & 0.84 & 1.10 & 29.07 & 0.12 \\
    {LINGO}~\cite{LINGO} & $\textcolor{red}{\usym{2717}}$ & 0.205 & \textbf{0.30} & 0.43 & 30.12 & 0.36 \\
    {CHOIS}~\cite{CHOIS} & $\textcolor{red}{\usym{2717}}$ & 0.111 & 0.64 & 1.17 & 39.72 & 0.09 \\
    \midrule
    {ZeroHSI}~\cite{ZeroHSI} & $\textcolor{green}{\usym{2713}}$ & 0.204 & 1.74 & 0.44 & 61.95 & 1.49 \\
    Ours & $\textcolor{green}{\usym{2713}}$ & \textbf{0.244} & 0.87 & \textbf{0.31} & \textbf{75.64} & \textbf{0.08} \\
    \bottomrule
  \end{tabular}
  \label{tab:interaction_metrics}
\end{table*}

\subsection{Interaction Quality Results}
\label{subsec:interaction_results}

\noindent \textbf{Quantitative Comparison.} Table~\ref{tab:interaction_metrics} presents the quantitative comparison of reconstruction quality metrics for articulated human-object scenes.
Our method demonstrates superior performance across multiple key areas.
\textbf{1)} We achieve the highest X-CLIP score ($0.244$), indicating superior semantic alignment between reconstructed interactions and textual descriptions.
\textbf{2)} In terms of foot sliding, our method achieves the lowest score ($0.31$), demonstrating more realistic foot contact during interactions.
\textbf{3)} Most notably, we achieve the highest contact percentage ($75.64\%$), showing that our method maintains more consistent human-object contact throughout the interaction sequence.
\textbf{4)} While non-zero-shot methods (TRUMANS: $0.84$, LINGO: $0.30$, CHOIS: $0.64$) achieve lower smoothness scores, these results should be interpreted with caution as \textit{false positives}.
The low smoothness scores of non-zero-shot methods stem from their minimal contact with articulated objects (Contact\%: TRUMANS $29.07\%$, LINGO $30.12\%$, CHOIS $39.72\%$), rather than genuine motion stability.
When human motion has limited interaction with objects, the smoothness metric can be artificially low due to the absence of complex contact dynamics that naturally introduce motion variations.
In contrast, our method maintains competitive smoothness ($0.87$) while achieving significantly higher contact rates ($75.64\%$), showing ArtHOI's ability to balance motion smoothness with realistic interaction complexity.
\textbf{5)} Our method achieves the lowest penetration errors ($0.08$), demonstrating superior physical plausibility compared to all baselines.
This pattern correlates with the contact results: our method achieves the highest contact percentage ($75.64\%$), while baselines show lower contact rates (ZeroHSI: $61.95\%$, CHOIS: $39.72\%$).
The superior penetration performance of our method demonstrates the effectiveness of our flow-based segmentation and two-stage optimization in maintaining physically plausible interactions with articulated objects.

\noindent \textbf{Qualitative Comparisons.}
\Cref{fig:qualitative_results} shows that our method recovers geometrically consistent and physically plausible 4D scenes with diverse articulated objects.  
Zero-shot 4D reconstruction baselines treat all objects as rigid entities, failing to model part-wise articulation.
%
%
%
In contrast, our method successfully recovers complex articulated interactions with superior consistency and physical plausibility by explicitly modeling object articulation through flow-based segmentation and two-stage decoupled reconstruction.

\subsection{Articulated Object Dynamics Results}
\label{subsec:articulation_results}

\begin{table}
  \centering
  
  \setlength{\tabcolsep}{8pt}
  \caption{Articulated object dynamics metrics under \textbf{monocular} setting (without multi-view input).}
  \begin{tabular}{l|ccc}
    \toprule
    Method & Rot (mean) ↓ & Rot (max) ↓ & Rot (min) ↓ \\
    \midrule
    {D3D-HOI}~\cite{xu2021d3dhoi} & 25.13 & 57.29 & 8.21 \\
    {3DADN}~\cite{qian2022understanding} & 21.17 & 55.21 & 5.62 \\
    \midrule
    Ours & \textbf{6.71} & \textbf{21.41} & \textbf{0.58} \\
    \bottomrule
  \end{tabular}
  \label{tab:articulation_metrics}
\end{table}

Table~\ref{tab:articulation_metrics} presents the comprehensive results for articulated object dynamics estimation.
Our method demonstrates dramatically superior performance across all metrics compared to specialized methods.
Our method achieves a mean rotation error of $6.71^\circ$, representing a $73.3\%$ reduction compared to D3D-HOI ($25.13^\circ$) and a $68.3\%$ reduction compared to 3DADN ($21.17^\circ$). Additionally, we achieve the lowest maximum rotation error ($21.41^\circ$ vs. $57.29^\circ$ / $55.21^\circ$) and minimum rotation error ($0.58^\circ$ vs. $8.21^\circ$ / $5.62^\circ$).
These results validate our core contribution: the ability to recover accurate articulated object dynamics from 2D video priors, enabling geometrically consistent and physically plausible 4D reconstruction without requiring 3D supervision.
The significant improvements in rotation estimation directly translate to more realistic and physically plausible articulated object motion during human-object interactions.

\subsection{Rigid Object Results}
\label{subsec:rigid}
Although ArtHOI is designed for articulated HOI, it naturally extends to rigid objects. Table~\ref{tab:rigid_comparison} compares our method with SAM3D+FP (SAM3D~\cite{sam3d}+FoundationPose~\cite{foundationpose}) and ZeroHSI on rigid object interactions under monocular RGB video. SAM3D+FP relies heavily on accurate depth priors and performs poorly with monocular input (low Contact\% $26.41\%$); ZeroHSI often produces weak human-object contact ($70.32\%$) and higher penetration ($1.52\%$). Our method achieves the best Foot Sliding ($0.28$), Contact\% ($76.18\%$), and Penetration\% ($0.06\%$), demonstrating that our reconstruction-informed synthesis generalizes to rigid objects while maintaining physical plausibility.

\begin{table}[t]
  \centering
  \small
  \setlength{\tabcolsep}{2pt}
  \caption{Rigid object comparison under monocular RGB.}
  \begin{tabular}{l|ccc}
    \toprule
    Method & Foot Sliding ↓ & Contact\% ↑ & Penetration\% ↓ \\
    \midrule
    SAM3D~\cite{sam3d}+FP~\cite{foundationpose} & 1.72 & 26.41 & 0.10 \\
    ZeroHSI~\cite{ZeroHSI} & 0.41 & 70.32 & 1.52 \\
    Ours & \textbf{0.28} & \textbf{76.18} & \textbf{0.06} \\
    \bottomrule
  \end{tabular}
  \label{tab:rigid_comparison}
\end{table}

\subsection{User Study}
\label{subsec:user_study}

To further validate the perceptual quality of our reconstructed 4D scenes, we conduct a comprehensive user study comparing our method with baseline approaches. The study involved $51$ participants with diverse backgrounds in computer graphics, robotics, and general technology. Each participant evaluated $20$ interaction sequences across different types of articulated objects (\eg, doors, cabinets, and fridges). Participants were presented with side-by-side comparisons of our method against each baseline (TRUMANS~\cite{TRUMANS}, CHOIS~\cite{CHOIS}, LINGO~\cite{LINGO}, ZeroHSI~\cite{ZeroHSI}) and asked to evaluate four criteria: \textbf{1) Realism}: how natural and physically plausible the human-object interactions appear; \textbf{2) Contact Quality}: the accuracy and consistency of contact between human body parts and articulated objects, including proper hand-object grasping; \textbf{3) Motion Smoothness}: the temporal consistency and fluidity of both human and object motion; \textbf{4) Overall Preference}: general preference ranking considering all aspects.

\noindent \textbf{Results.} Table~\ref{tab:user_study} presents the comprehensive user study results across all evaluation dimensions. Our method demonstrates superior performance compared to all baseline approaches, with participants consistently preferring our reconstructed interactions. Specifically, our method achieves the highest preference rates against TRUMANS ($98.04\%$ overall), CHOIS ($95.28\%$ overall), LINGO ($91.51\%$ overall), and ZeroHSI ($89.42\%$ overall). The results particularly highlight our method's strength in \textbf{Contact Quality} and \textbf{Motion Smoothness}, where we achieve $98.00\%$ and $92.16\%$ preference rates against TRUMANS, respectively. This validates that our flow-based segmentation and two-stage reconstruction effectively capture the complex dynamics of articulated human-object interactions.

\begin{figure*}[t]
    \centering
    \includegraphics[width=0.97\linewidth]{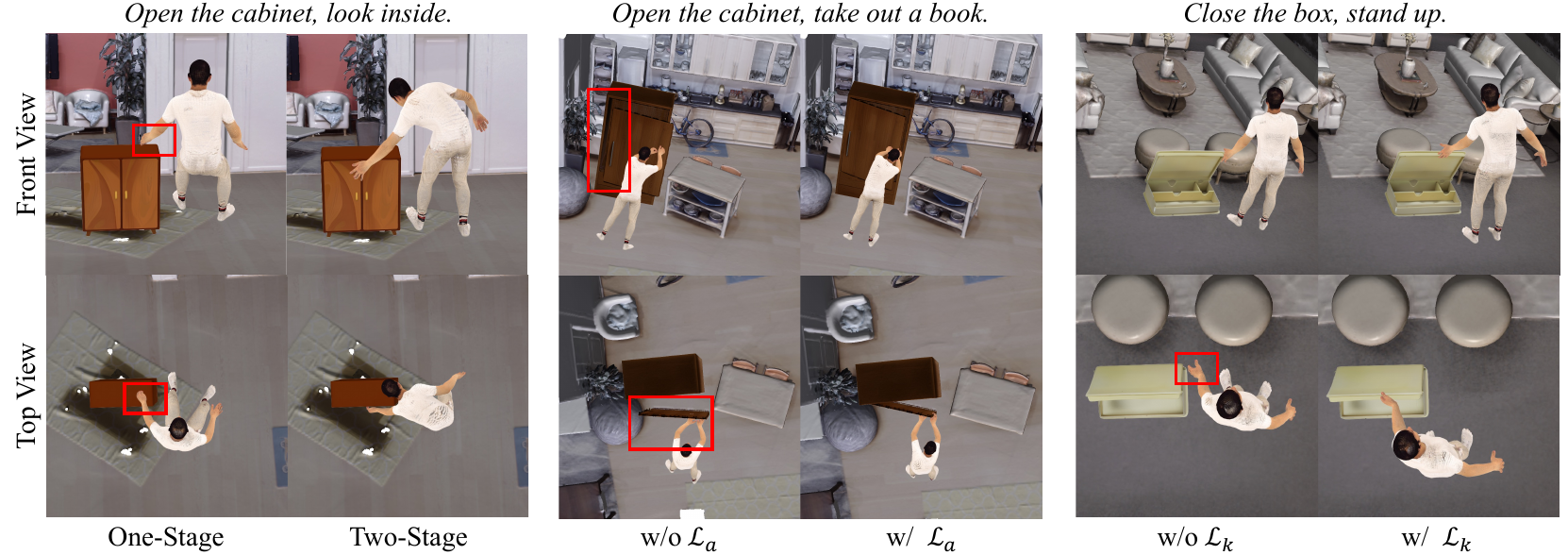}
    \caption{Comparing our full model with variants. Better inspected in our supplementary video.}
    \label{fig:ablation}
\end{figure*}
\subsection{Ablation Studies}
\label{subsec:ablation}
We conduct extensive ablations to analyze the contribution of each key component in our framework.
Table~\ref{tab:ablation} presents the quantitative results when removing individual components. We evaluate on both interaction quality (X-CLIP, Foot Sliding, Contact\%) and articulated object dynamics (rotation errors).

\noindent \textbf{Two-stage decoupling vs. joint optimization.}
Replacing our two-stage pipeline with joint optimization (optimizing object articulation and human motion simultaneously) yields the largest degradation across interaction metrics.
X-CLIP drops from $0.244$ to $0.187$, and Contact\% falls from $75.64\%$ to $61.45\%$; rotation errors also increase substantially (Rot mean: $12.34^\circ$ vs. $6.71^\circ$).
As shown in~\Cref{fig:ablation} (a), joint optimization fails to learn geometrically consistent interactions because gradients from reconstruction, kinematic, and contact terms compete across coupled human-object parameters, leading to unstable convergence and inconsistent object articulation.
Our two-stage design first recovers a stable 4D object scaffold (Stage I) and then refines human motion under fixed geometry (Stage II), avoiding this interference and enabling accurate contact alignment.

\noindent \textbf{Articulation regularization $\mathcal{L}_a$.}
Removing $\mathcal{L}_a$ (distance preservation between quasi-static binding pairs) causes the largest increase in articulation error: Rot (mean) rises from $6.71^\circ$ to $15.67^\circ$ (a $133\%$ increase), and Rot (max) from $21.41^\circ$ to $42.18^\circ$.
Without $\mathcal{L}_a$, the dynamic and static parts of articulated objects are no longer constrained to maintain rigid-body relationships at the hinge; articulated parts drift away from the main body, violating physical constraints.
As illustrated in~\Cref{fig:ablation} (b), this leads to unrealistic object configurations and, consequently, degraded interaction quality (Contact\% $68.75\%$ vs. $75.64\%$) because the 3D scaffold for hand-object contact becomes geometrically incorrect.

\noindent \textbf{Kinematic loss $\mathcal{L}_k$.}
Removing the kinematic loss $\mathcal{L}_k$ (which pulls hand joints toward 3D contact keypoints derived from Stage I) has a dramatic effect on interaction metrics despite leaving articulation errors unchanged.
Contact\% drops from $75.64\%$ to $59.82\%$ (the lowest among all ablations), X-CLIP from $0.244$ to $0.201$, and Foot Sliding worsens from $0.31$ to $0.58$.
This confirms that $\mathcal{L}_k$ is essential for aligning human motion with the reconstructed object geometry.
Without it, the optimization relies only on 2D reconstruction and prior terms, which are insufficient to resolve monocular depth ambiguity; hands drift away from the object surface, leading to weak contact and higher foot sliding as the optimizer compensates with unstable pose changes.
As shown in~\Cref{fig:ablation} (c), removing $\mathcal{L}_k$ results in visibly misaligned hand-object contact.
Thus, the 3D contact keypoints derived from our flow-based segmentation and Stage I reconstruction directly govern the plausibility of human-object interaction.

\noindent \textbf{Smoothness loss $\mathcal{L}_s$.}
Removing $\mathcal{L}_s$ degrades both articulation (Rot mean: $8.23^\circ$, Rot max: $25.45^\circ$) and interaction (Contact\% $65.43\%$, Foot Sliding $0.49$).
$\mathcal{L}_s$ penalizes abrupt changes in articulation transforms and pose parameters across frames, promoting temporally coherent motion trajectories.
Without it, optimization tends to overfit per-frame 2D cues, producing jittery articulations and unstable hand trajectories that reduce contact consistency and increase rotation variance.

\begin{table*}[!t]
  \centering
  
  \setlength{\tabcolsep}{19pt}
  \caption{User study shows the percentage of participants who preferred our method over each baseline across four evaluation dimensions.}
  \begin{tabular}{l|cccc}
    \toprule
    Method & Realism ↑ & Contact Quality ↑ & Motion Smoothness ↑ & Overall Preference ↑ \\
    \midrule
    Ours vs. {TRUMANS}~\cite{TRUMANS} & 96.08\% & 98.00\% & 92.16\% & 98.04\% \\
    Ours vs. {CHOIS}~\cite{CHOIS} & 95.20\% & 89.08\% & 94.83\% & 95.28\% \\
    Ours vs. {LINGO}~\cite{LINGO} & 90.20\% & 87.13\% & 92.00\% & 91.51\% \\
    Ours vs. {ZeroHSI}~\cite{ZeroHSI} & 91.18\% & 85.41\% & 84.21\% & 89.42\% \\
    \bottomrule
  \end{tabular}
  \label{tab:user_study}
\end{table*}

\begin{table*}[!t]
  \centering
  
  \setlength{\tabcolsep}{16pt}
  \caption{Ablation study results. We remove individual components and evaluate their impact on both interaction and articulation.}
  \begin{tabular}{l|ccc|ccc}
    \toprule
    \multirow{2}{*}{\textbf{Method}} & \multicolumn{3}{c|}{\textbf{Interaction}} & \multicolumn{3}{c}{\textbf{Articulation}} \\
    & X-CLIP ↑ & Foot Sliding ↓ & Contact\% ↑ & Rot (mean) ↓ & Rot (max) ↓ & Rot (min) ↓ \\
    \midrule
    Joint Opt. & 0.187 & 0.67 & 61.45 & 12.34 & 35.89 & 2.01 \\
    w/o $\mathcal{L}_a$ & 0.223 & 0.42 & 68.75 & 15.67 & 42.18 & 4.56 \\
    w/o $\mathcal{L}_k$ & 0.201 & 0.58 & 59.82 & \bf 6.71 & \bf 21.41 & \bf 0.58 \\
    w/o $\mathcal{L}_s$ & 0.218 & 0.49 & 65.43 & 8.23 & 25.45 & 0.79 \\
    \midrule
    Full Model & \bf 0.244 & \bf 0.31 & \bf 75.64 & \bf 6.71 & \bf 21.41 & \bf 0.58 \\
    \bottomrule
  \end{tabular}
  \label{tab:ablation}
\end{table*}

%% file: sections/5_conclusion.tex
\section{Discussion and Conclusion}
\label{sec:conclusion}

\subsection{Conclusion}
We present \methodname, the first zero-shot framework for synthesizing articulated human-object interactions by reconstructing full 4D scenes from monocular video priors. Existing zero-shot methods are inherently restricted to rigid objects and thus cannot handle everyday articulated objects such as doors, drawers, and cabinets. Our framework addresses this limitation and produces temporally coherent 4D mesh sequences with explicit part articulation and contact.
Our key insight is to formulate interaction synthesis as a 4D reconstruction problem. Rather than generating interactions end-to-end through differentiable rendering, we synthesize 3D interactions by reconstructing full 4D articulated scenes from generated 2D videos. We use flow-guided part segmentation to decompose the object into rigid parts and decoupled optimization to recover poses and articulation from 2D observations, effectively solving an inverse rendering problem.
This reconstruction-based design injects explicit geometric and kinematic priors into monocular dynamics. As a result, the synthesized interactions are both semantically aligned with the intended action and physically plausible in terms of part motion and contact.
Extensive experiments on standard benchmarks show that our method achieves superior geometric consistency, contact accuracy, and temporal coherence compared with prior zero-shot approaches, while remaining efficient enough for practical use.

\subsection{Limitation}
\begin{figure}[t]
  \centering
  \includegraphics[width=\columnwidth]{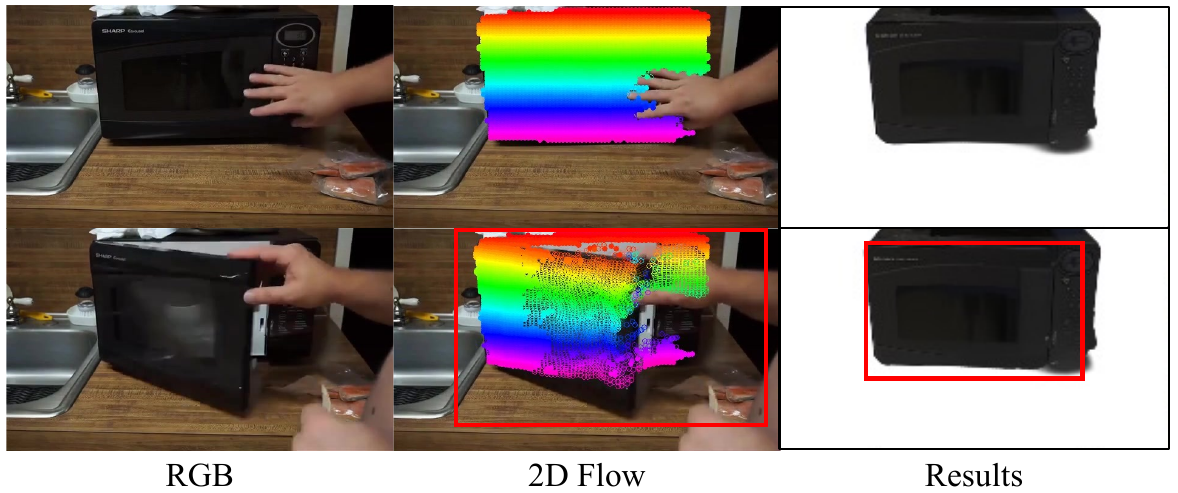}
  \caption{Failure cases. Co-tracker struggles with low-texture or reflective regions, leading to distortions that propagate into articulation prediction.}
  \label{fig:failcase}
\end{figure}

Our experiments focus on single-part articulated objects, which form the basic blocks of hierarchical articulated structures. Fig.~\ref{fig:failcase} illustrates representative failure cases.
\textbf{1) Optical flow tracking failures.} Co-tracker struggles with low-texture or reflective regions, leading to distortions that propagate into articulation prediction. When articulated object surfaces lack sufficient visual features or contain specular reflections, optical flow becomes unreliable and flow-based segmentation fails.
\textbf{2) Complex articulated structures.} Our method struggles with objects having multiple DOF or non-rigid articulations (\eg, soft-body joints, elastic connections).
\textbf{3) Long-term temporal consistency.} As sequences lengthen, cumulative errors in articulation extraction can lead to gradual deviation from physical plausibility.
\textbf{4) Fixed camera assumption.} We assume fixed cameras. Moving cameras introduce severe mixed ego-motion and object articulation, making kinematic recovery much harder.

\subsection{Broader Impact}
Our work has practical implications across multiple domains. In \textbf{robotics}, \methodname can generate training data for manipulation policies involving articulated objects without expensive motion capture or manual annotation, enabling scalable simulation-to-real transfer. In \textbf{virtual and augmented reality}, it supports the creation of realistic human-object interactions for games, virtual environments, and telepresence applications, reducing the need for hand-crafted animations. In \textbf{embodied AI and data synthesis}, the zero-shot pipeline allows rapid generation of diverse, physically plausible 4D interaction datasets from text prompts, facilitating research in scene understanding, action recognition, and contact-aware motion prediction. The efficiency of our approach (approximately $30$ minutes per scene on a single GPU) makes it suitable for rapid prototyping and iterative content creation. 

%% file: main.bib
@string{Optimization = {Optimization} }

@String{ACMTG = "{ACM} Trans. Graph." }

@String{SIGGRAPH = "Spec. Interest Group Comput. Graph. Interactive Technol." }

@String{Computer = "{IEEE} Computer" }

@String{CVFIEEE = "Computer Vision Foundation / {IEEE}" }

@string{ICCV = {ICCV} }

@string{ECCV = {ECCV} }

@string{CVPR = {CVPR} }

@string{NIPS = {NIPS} }

@string{ACMMM = {ACMMM} }

@string{ICLR = {ICLR}}

@string{WACV = {WACV} }

@string{CoRR = {arXiv Comput. Res. Repository}}

@article{SMPL,
  author    = {Matthew Loper and
               Naureen Mahmood and
               Javier Romero and
               Gerard Pons{-}Moll and
               Michael J. Black},
  title     = {{SMPL:} a skinned multi-person linear model},
  journal   = ACMTG,
  volume    = {34},
  number    = {6},
  pages     = {248:1--248:16},
  year      = {2015},

  doi       = {10.1145/2816795.2818013},
  bibsource = {dblp computer science bibliography, https://dblp.org}
}

@inproceedings{SMPLX,
  author    = {Georgios Pavlakos and
               Vasileios Choutas and
               Nima Ghorbani and
               Timo Bolkart and
               Ahmed A. A. Osman and
               Dimitrios Tzionas and
               Michael J. Black},
  title     = {Expressive Body Capture: 3D Hands, Face, and Body From a Single Image},
  booktitle = CVPR,
  pages     = {10975--10985},
  publisher = CVFIEEE,
  year      = {2019},

  doi       = {10.1109/CVPR.2019.01123},
  bibsource = {dblp computer science bibliography, https://dblp.org}
}

@inproceedings{smplify,
  author    = {Federica Bogo and
               Angjoo Kanazawa and
               Christoph Lassner and
               Peter V. Gehler and
               Javier Romero and
               Michael J. Black},
  editor    = {Bastian Leibe and
               Jiri Matas and
               Nicu Sebe and
               Max Welling},
  title     = {Keep It {SMPL:} Automatic Estimation of 3D Human Pose and Shape from
               a Single Image},
  booktitle = ECCV,
  series    = {Lecture Notes in Computer Science},
  volume    = {9909},
  pages     = {561--578},
  year      = {2016},

  doi       = {10.1007/978-3-319-46454-1\_34},
  bibsource = {dblp computer science bibliography, https://dblp.org}
}

@inproceedings{Adam,
  title={Adam: A Method for Stochastic Optimization},
  author={Diederik P. Kingma and Jimmy Ba},
  booktitle=ICLR,
  year={2014},
  pages={},
  doi={}
}

@inproceedings{pytorch,
 author = {Paszke, Adam and Gross, Sam and Massa, Francisco and Lerer, Adam and Bradbury, James and Chanan, Gregory and Killeen, Trevor and Lin, Zeming and Gimelshein, Natalia and Antiga, Luca and Desmaison, Alban and Kopf, Andreas and Yang, Edward and DeVito, Zachary and Raison, Martin and Tejani, Alykhan and Chilamkurthy, Sasank and Steiner, Benoit and Fang, Lu and Bai, Junjie and Chintala, Soumith},
 booktitle = NIPS,
 pages = {},

 address = {},
 title = {PyTorch: An Imperative Style, High-Performance Deep Learning Library},
 volume = {32},
 year = {2019}
}

@article{SAM,
  title={Segment Anything},
  author={Kirillov, Alexander and Mintun, Eric and Ravi, Nikhila and Mao, Hanzi and Rolland, Chloe and Gustafson, Laura and Xiao, Tete and Whitehead, Spencer and Berg, Alexander C. and Lo, Wan-Yen and Doll{\'a}r, Piotr and Girshick, Ross},
  journal={arXiv:2304.02643},
  year={2023}
}

@article{SAM2,
  title={Sam 2: Segment anything in images and videos},
  author={Ravi, Nikhila and Gabeur, Valentin and Hu, Yuan-Ting and Hu, Ronghang and Ryali, Chaitanya and Ma, Tengyu and Khedr, Haitham and R{\"a}dle, Roman and Rolland, Chloe and Gustafson, Laura and others},
  journal={arXiv:2408.00714},
  year={2024}
}

@inproceedings{co-tracker,
  title={Cotracker: It is better to track together},
  author={Karaev, Nikita and Rocco, Ignacio and Graham, Benjamin and Neverova, Natalia and Vedaldi, Andrea and Rupprecht, Christian},
  booktitle=ECCV,
  pages={18--35},
  year={2024},

}

@inproceedings{gvhmr,
  title={World-Grounded Human Motion Recovery via Gravity-View Coordinates},
  author={Shen, Zehong and Pi, Huaijin and Xia, Yan and Cen, Zhi and Peng, Sida and Hu, Zechen and Bao, Hujun and Hu, Ruizhen and Zhou, Xiaowei},
  booktitle={SIGGRAPH Asia Conference Proceedings},
  year={2024}
}

@inproceedings{TRUMANS,
  title={Scaling up dynamic human-scene interaction modeling},
  author={Jiang, Nan and Zhang, Zhiyuan and Li, Hongjie and Ma, Xiaoxuan and Wang, Zan and Chen, Yixin and Liu, Tengyu and Zhu, Yixin and Huang, Siyuan},
  booktitle=CVPR,
  pages={1737--1747},
  year={2024}
}

@article{ZeroHSI,
  title={Zerohsi: Zero-shot 4d human-scene interaction by video generation},
  author={Li, Hongjie and Yu, Hong-Xing and Li, Jiaman and Wu, Jiajun},
  journal={arXiv:2412.18600},
  year={2024}
}

@inproceedings{LINGO,
  title={Autonomous character-scene interaction synthesis from text instruction},
  author={Jiang, Nan and He, Zimo and Wang, Zi and Li, Hongjie and Chen, Yixin and Huang, Siyuan and Zhu, Yixin},
  booktitle={SIGGRAPH Asia},
  pages={1--11},
  year={2024}
}

@inproceedings{CHOIS,
  title={Controllable human-object interaction synthesis},
  author={Li, Jiaman and Clegg, Alexander and Mottaghi, Roozbeh and Wu, Jiajun and Puig, Xavier and Liu, C Karen},
  booktitle=ECCV,
  pages={54--72},
  year={2024},

}

@inproceedings{Nifty,
  title={Nifty: Neural object interaction fields for guided human motion synthesis},
  author={Kulkarni, Nilesh and Rempe, Davis and Genova, Kyle and Kundu, Abhijit and Johnson, Justin and Fouhey, David and Guibas, Leonidas},
  booktitle=CVPR,
  pages={947--957},
  year={2024}
}

@inproceedings{GenZi,
  title={Genzi: Zero-shot 3d human-scene interaction generation},
  author={Li, Lei and Dai, Angela},
  booktitle=CVPR,
  pages={20465--20474},
  year={2024}
}

@article{InterDreamer,
  title={Interdreamer: Zero-shot text to 3d dynamic human-object interaction},
  author={Xu, Sirui and Wang, Yu-Xiong and Gui, Liangyan and others},
  journal=NIPS,
  volume={37},
  pages={52858--52890},
  year={2024}
}

@article{li2023object,
  title={Object motion guided human motion synthesis},
  author={Li, Jiaman and Wu, Jiajun and Liu, C Karen},
  journal={ACM Transactions on Graphics (TOG)},
  volume={42},
  number={6},
  pages={1--11},
  year={2023},

}

@inproceedings{gao2020interactgan,
  title={Interactgan: Learning to generate human-object interaction},
  author={Gao, Chen and Liu, Si and Zhu, Defa and Liu, Quan and Cao, Jie and He, Haoqian and He, Ran and Yan, Shuicheng},
  booktitle=ACMMM,
  pages={165--173},
  year={2020}
}

@inproceedings{diller2024cg,
  title={Cg-hoi: Contact-guided 3d human-object interaction generation},
  author={Diller, Christian and Dai, Angela},
  booktitle=CVPR,
  pages={19888--19901},
  year={2024}
}

@inproceedings{li2024task,
  title={Task-oriented human-object interactions generation with implicit neural representations},
  author={Li, Quanzhou and Wang, Jingbo and Loy, Chen Change and Dai, Bo},
  booktitle=WACV,
  pages={3035--3044},
  year={2024}
}

@inproceedings{xu2023interdiff,
  title={Interdiff: Generating 3d human-object interactions with physics-informed diffusion},
  author={Xu, Sirui and Li, Zhengyuan and Wang, Yu-Xiong and Gui, Liang-Yan},
  booktitle=ICCV,
  pages={14928--14940},
  year={2023}
}

@article{fan20253d,
  title={3D Human Interaction Generation: A Survey},
  author={Fan, Siyuan and Huang, Wenke and Cai, Xiantao and Du, Bo},
  journal={arXiv:2503.13120},
  year={2025}
}

@article{huang2025hunyuanvideo,
  title={HunyuanVideo-HOMA: Generic Human-Object Interaction in Multimodal Driven Human Animation},
  author={Huang, Ziyao and Zhou, Zixiang and Cao, Juan and Ma, Yifeng and Chen, Yi and Rao, Zejing and Xu, Zhiyong and Wang, Hongmei and Lin, Qin and Zhou, Yuan and others},
  journal={arXiv:2506.08797},
  year={2025}
}

@inproceedings{zhang2025interactanything,
  title={InteractAnything: Zero-shot Human Object Interaction Synthesis via LLM Feedback and Object Affordance Parsing},
  author={Zhang, Jinlu and Chen, Yixin and Wang, Zan and Yang, Jie and Wang, Yizhou and Huang, Siyuan},
  booktitle=CVPR,
  pages={7015--7025},
  year={2025}
}

@article{li2025genhoi,
  title={GenHOI: Generalizing Text-driven 4D Human-Object Interaction Synthesis for Unseen Objects},
  author={Li, Shujia and Zhang, Haiyu and Chen, Xinyuan and Wang, Yaohui and Ban, Yutong},
  journal={arXiv:2506.15483},
  year={2025}
}

@inproceedings{jiang2023full,
  title={Full-body articulated human-object interaction},
  author={Jiang, Nan and Liu, Tengyu and Cao, Zhexuan and Cui, Jieming and Zhang, Zhiyuan and Chen, Yixin and Wang, He and Zhu, Yixin and Huang, Siyuan},
  booktitle=ICCV,
  pages={9365--9376},
  year={2023}
}

@article{chao2025part,
  title={Part Segmentation and Motion Estimation for Articulated Objects with Dynamic 3D Gaussians},
  author={Chao, Jun-Jee and Jiang, Qingyuan and Isler, Volkan},
  journal={arXiv:2506.22718},
  year={2025}
}

@inproceedings{song2024reacto,
  title={Reacto: Reconstructing articulated objects from a single video},
  author={Song, Chaoyue and Wei, Jiacheng and Foo, Chuan Sheng and Lin, Guosheng and Liu, Fayao},
  booktitle=CVPR,
  pages={5384--5395},
  year={2024}
}

@article{deng2024articulate,
  title={Articulate your NeRF: Unsupervised articulated object modeling via conditional view synthesis},
  author={Deng, Jianning and Subr, Kartic and Bilen, Hakan},
  journal=NIPS,
  volume={37},
  pages={119717--119741},
  year={2024}
}

@article{peng2025generalizable,
  title={Generalizable Articulated Object Reconstruction from Casually Captured RGBD Videos},
  author={Peng, Weikun and Lv, Jun and Lu, Cewu and Savva, Manolis},
  journal={arXiv:2506.08334},
  year={2025}
}

@article{xu2021articulated,
  title={Articulated object reconstruction from interaction videos},
  author={Xu, Xiang},
  year={2021},

}

@article{3dgs,
  title={3D Gaussian splatting for real-time radiance field rendering.},
  author={Kerbl, Bernhard and Kopanas, Georgios and Leimk{\"u}hler, Thomas and Drettakis, George},
  journal={ACM Trans. Graph.},
  volume={42},
  number={4},
  pages={139--1},
  year={2023}
}

@article{xu2021d3dhoi,
  title={D3D-HOI: Dynamic 3D Human-Object Interactions from Videos},
  author={Xiang Xu and Hanbyul Joo and Greg Mori and Manolis Savva},
  journal={arXiv:2108.08420},
  year={2021}
}

@inproceedings{qian2022understanding,
  title={Understanding 3d object articulation in internet videos},
  author={Qian, Shengyi and Jin, Linyi and Rockwell, Chris and Chen, Siyi and Fouhey, David F},
  booktitle=CVPR,
  pages={1599--1609},
  year={2022}
}

@inproceedings{shen2023xavatar,
      title={X-Avatar: Expressive Human Avatars},
      author={Shen, Kaiyue and Guo, Chen and Kaufmann, Manuel and Zarate, Juan and Valentin, Julien and Song, Jie and Hilliges, Otmar},    
      journal   = {Computer Vision and Pattern Recognition (CVPR)},
      year      = {2023}
}

@article{trellis,
    title   = {Structured 3D Latents for Scalable and Versatile 3D Generation},
    author  = {Xiang, Jianfeng and Lv, Zelong and Xu, Sicheng and Deng, Yu and Wang, Ruicheng and Zhang, Bowen and Chen, Dong and Tong, Xin and Yang, Jiaolong},
    journal = {arXiv:2412.01506},
    year    = {2024}
}

@article{straub2019replica,
  title={The replica dataset: A digital replica of indoor spaces},
  author={Straub, Julian and Whelan, Thomas and Ma, Lingni and Chen, Yufan and Wijmans, Erik and Green, Simon and Engel, Jakob J and Mur-Artal, Raul and Ren, Carl and Verma, Shobhit and others},
  journal={arXiv:1906.05797},
  year={2019}
}

@inproceedings{xclip,
  title={Expanding language-image pretrained models for general video recognition},
  author={Ni, Bolin and Peng, Houwen and Chen, Minghao and Zhang, Songyang and Meng, Gaofeng and Fu, Jianlong and Xiang, Shiming and Ling, Haibin},
  booktitle=ECCV,
  pages={1--18},
  year={2022},

}

@article{park2025zero4d,
  title={Zero4D: Training-Free 4D Video Generation From Single Video Using Off-the-Shelf Video Diffusion Model},
  author={Park, Jangho and Kwon, Taesung and Ye, Jong Chul},
  journal={arXiv e-prints},
  pages={arXiv--2503},
  year={2025}
}

@inproceedings{wang2025videoscene,
  title={VideoScene: Distilling video diffusion model to generate 3D scenes in one step},
  author={Wang, Hanyang and Liu, Fangfu and Chi, Jiawei and Duan, Yueqi},
  booktitle=CVPR,
  pages={16475--16485},
  year={2025},

}

@article{liu2025free4d,
  title={Free4D: Tuning-free 4D Scene Generation with Spatial-Temporal Consistency},
  author={Liu, Tianqi and Huang, Zihao and Chen, Zhaoxi and Wang, Guangcong and Hu, Shoukang and Shen, Liao and Sun, Huiqiang and Cao, Zhiguo and Li, Wei and Liu, Ziwei},
  journal={arXiv:2503.20785},
  year={2025}
}

@inproceedings{liu2025building,
  title={Building interactable replicas of complex articulated objects via gaussian splatting},
  author={Liu, Yu and Jia, Baoxiong and Lu, Ruijie and Ni, Junfeng and Zhu, Song-Chun and Huang, Siyuan},
  booktitle=ICLR,
  year={2025}
}

@inproceedings{su2025artformer,
  title={Artformer: Controllable generation of diverse 3d articulated objects},
  author={Su, Jiayi and Feng, Youhe and Li, Zheng and Song, Jinhua and He, Yangfan and Ren, Botao and Xu, Botian},
  booktitle=CVPR,
  pages={1894--1904},
  year={2025}
}

@inproceedings{zhai2025taga,
  title={TAGA: Self-supervised Learning for Template-free Animatable Gaussian Articulated Model},
  author={Zhai, Zhichao and Chen, Guikun and Wang, Wenguan and Zheng, Dong and Xiao, Jun},
  booktitle=CVPR,
  pages={21159--21169},
  year={2025}
}

@inproceedings{yao2025riggs,
  title={Riggs: Rigging of 3d gaussians for modeling articulated objects in videos},
  author={Yao, Yuxin and Deng, Zhi and Hou, Junhui},
  booktitle=CVPR,
  pages={5592--5601},
  year={2025}
}

@inproceedings{gao2025meshart,
  title={MeshArt: Generating Articulated Meshes with Structure-guided Transformers},
  author={Gao, Daoyi and Siddiqui, Yawar and Li, Lei and Dai, Angela},
  booktitle=CVPR,
  pages={618--627},
  year={2025}
}

@inproceedings{guo2025articulatedgs,
  title={Articulatedgs: Self-supervised digital twin modeling of articulated objects using 3d gaussian splatting},
  author={Guo, Junfu and Xin, Yu and Liu, Gaoyi and Xu, Kai and Liu, Ligang and Hu, Ruizhen},
  booktitle=CVPR,
  pages={27144--27153},
  year={2025}
}

@inproceedings{li2025articulated,
  title={Articulated Kinematics Distillation from Video Diffusion Models},
  author={Li, Xuan and Ma, Qianli and Lin, Tsung-Yi and Chen, Yongxin and Jiang, Chenfanfu and Liu, Ming-Yu and Xiang, Donglai},
  booktitle=CVPR,
  pages={17571--17581},
  year={2025}
}

@article{lin2025splart,
  title={SplArt: Articulation Estimation and Part-Level Reconstruction with 3D Gaussian Splatting},
  author={Lin, Shengjie and Fang, Jiading and Irshad, Muhammad Zubair and Guizilini, Vitor Campagnolo and Ambrus, Rares Andrei and Shakhnarovich, Greg and Walter, Matthew R},
  journal={arXiv:2506.03594},
  year={2025}
}

@article{goyal2025geopard,
  title={GEOPARD: Geometric Pretraining for Articulation Prediction in 3D Shapes},
  author={Goyal, Pradyumn and Petrov, Dmitry and Andrews, Sheldon and Ben-Shabat, Yizhak and Liu, Hsueh-Ti Derek and Kalogerakis, Evangelos},
  journal={arXiv:2504.02747},
  year={2025}
}

@article{zhang2025adaptive,
  title={Adaptive Articulated Object Manipulation On The Fly with Foundation Model Reasoning and Part Grounding},
  author={Zhang, Xiaojie and Wang, Yuanfei and Wu, Ruihai and Xu, Kunqi and Li, Yu and Xiang, Liuyu and Dong, Hao and He, Zhaofeng},
  journal={arXiv:2507.18276},
  year={2025}
}

@article{kreber2025guiding,
  title={Guiding diffusion-based articulated object generation by partial point cloud alignment and physical plausibility constraints},
  author={Kreber, Jens U and Stueckler, Joerg},
  journal={arXiv:2508.00558},
  year={2025}
}

@article{zhang2025physrig,
  title={PhysRig: Differentiable Physics-Based Skinning and Rigging Framework for Realistic Articulated Object Modeling},
  author={Zhang, Hao and Xu, Haolan and Feng, Chun and Jampani, Varun and Ahuja, Narendra},
  journal={arXiv:2506.20936},
  year={2025}
}

@inproceedings{behave,
  author       = {Bharat Lal Bhatnagar and
                  Xianghui Xie and
                  Ilya A. Petrov and
                  Cristian Sminchisescu and
                  Christian Theobalt and
                  Gerard Pons{-}Moll},
  title        = {{BEHAVE:} Dataset and Method for Tracking Human Object Interactions},
  booktitle    = CVPR,
  pages        = {15914--15925},

  year         = {2022},

  doi          = {10.1109/CVPR52688.2022.01547},
  bibsource    = {dblp computer science bibliography, https://dblp.org}
}

@inproceedings{foundationpose,
  author       = {Bowen Wen and
                  Wei Yang and
                  Jan Kautz and
                  Stan Birchfield},
  title        = {FoundationPose: Unified 6D Pose Estimation and Tracking of Novel Objects},
  booktitle    = CVPR,
  pages        = {17868--17879},

  year         = {2024},

  doi          = {10.1109/CVPR52733.2024.01692},
  bibsource    = {dblp computer science bibliography, https://dblp.org}
}

@inproceedings{Stereo4D,
  author       = {Linyi Jin and
                  Richard Tucker and
                  Zhengqi Li and
                  David Fouhey and
                  Noah Snavely and
                  Aleksander Holynski},
  title        = {Stereo4D: Learning How Things Move in 3D from Internet Stereo Videos},
  booktitle    = CVPR,
  pages        = {10497--10509},

  year         = {2025},

  doi          = {10.1109/CVPR52734.2025.00982},
  bibsource    = {dblp computer science bibliography, https://dblp.org}
}

@inproceedings{DNF,
  author       = {Xinyi Zhang and
                  Naiqi Li and
                  Angela Dai},
  title        = {{DNF:} Unconditional 4D Generation with Dictionary-based Neural Fields},
  booktitle    = CVPR,
  pages        = {26047--26056},

  year         = {2025},

  doi          = {10.1109/CVPR52734.2025.02426},
  bibsource    = {dblp computer science bibliography, https://dblp.org}
}

@inproceedings{DBLP:conf/cvpr/KwonC025,
  author       = {JooHyun Kwon and
                  Hanbyel Cho and
                  Junmo Kim},
  title        = {Efficient Dynamic Scene Editing via 4D Gaussian-based Static-Dynamic
                  Separation},
  booktitle    = CVPR,
  pages        = {26855--26865},

  year         = {2025},

  doi          = {10.1109/CVPR52734.2025.02501},
  bibsource    = {dblp computer science bibliography, https://dblp.org}
}

@inproceedings{DIO,
  author       = {Christopher Diehl and
                  Quinlan Sykora and
                  Ben Agro and
                  Thomas Gilles and
                  Sergio Casas and
                  Raquel Urtasun},
  title        = {{DIO:} Decomposable Implicit 4D Occupancy-Flow World Model},
  booktitle    = CVPR,
  pages        = {27456--27466},

  year         = {2025},

  doi          = {10.1109/CVPR52734.2025.02557},
  bibsource    = {dblp computer science bibliography, https://dblp.org}
}

@inproceedings{4DLangSplat,
  author       = {Wanhua Li and
                  Renping Zhou and
                  Jiawei Zhou and
                  Yingwei Song and
                  Johannes Herter and
                  Minghan Qin and
                  Gao Huang and
                  Hanspeter Pfister},
  title        = {4D LangSplat: 4D Language Gaussian Splatting via Multimodal Large
                  Language Models},
  booktitle    = CVPR,
  pages        = {22001--22011},

  year         = {2025},
  doi          = {10.1109/CVPR52734.2025.02049},
  bibsource    = {dblp computer science bibliography, https://dblp.org}
}

@inproceedings{Mamba4D,
  author       = {Jiuming Liu and
                  Jinru Han and
                  Lihao Liu and
                  Angelica I. Avil{\'{e}}s{-}Rivero and
                  Chaokang Jiang and
                  Zhe Liu and
                  Hesheng Wang},
  title        = {Mamba4D: Efficient 4D Point Cloud Video Understanding with Disentangled
                  Spatial-Temporal State Space Models},
  booktitle    = CVPR,
  pages        = {17626--17636},

  year         = {2025},

  doi          = {10.1109/CVPR52734.2025.01642},
  bibsource    = {dblp computer science bibliography, https://dblp.org}
}

@inproceedings{Feature4X,
  author       = {Shijie Zhou and
                  Hui Ren and
                  Yijia Weng and
                  Shuwang Zhang and
                  Zhen Wang and
                  Dejia Xu and
                  Zhiwen Fan and
                  Suya You and
                  Zhangyang Wang and
                  Leonidas J. Guibas and
                  Achuta Kadambi},
  title        = {Feature4X: Bridging Any Monocular Video to 4D Agentic {AI} with Versatile
                  Gaussian Feature Fields},
  booktitle    = CVPR,
  pages        = {14179--14190},

  year         = {2025},

  doi          = {10.1109/CVPR52734.2025.01323},
  bibsource    = {dblp computer science bibliography, https://dblp.org}
}

@inproceedings{DBLP:conf/cvpr/TaubnerZTL25,
  author       = {Felix Taubner and
                  Ruihang Zhang and
                  Mathieu Tuli and
                  David B. Lindell},
  title        = {{CAP4D:} Creating Animatable 4D Portrait Avatars with Morphable Multi-View
                  Diffusion Models},
  booktitle    = CVPR,
  pages        = {5318--5330},

  year         = {2025},

  doi          = {10.1109/CVPR52734.2025.00501},
  bibsource    = {dblp computer science bibliography, https://dblp.org}
}

@inproceedings{DBLP:conf/cvpr/LiuZXTYY25,
  author       = {Yun Liu and
                  Chengwen Zhang and
                  Ruofan Xing and
                  Bingda Tang and
                  Bowen Yang and
                  Li Yi},
  title        = {{CORE4D:} {A} 4D Human-Object-Human Interaction Dataset for Collaborative
                  Object REarrangement},
  booktitle    = CVPR,
  pages        = {1769--1782},

  year         = {2025},

  doi          = {10.1109/CVPR52734.2025.00172},
  bibsource    = {dblp computer science bibliography, https://dblp.org}
}

@inproceedings{DBLP:conf/cvpr/ZhaoNWZZW0CWZMW25,
  author       = {Guosheng Zhao and
                  Chaojun Ni and
                  Xiaofeng Wang and
                  Zheng Zhu and
                  Xueyang Zhang and
                  Yida Wang and
                  Guan Huang and
                  Xinze Chen and
                  Boyuan Wang and
                  Youyi Zhang and
                  Wenjun Mei and
                  Xingang Wang},
  title        = {DriveDreamer4D: World Models Are Effective Data Machines for 4D Driving
                  Scene Representation},
  booktitle    = CVPR,
  pages        = {12015--12026},

  year         = {2025},

  doi          = {10.1109/CVPR52734.2025.01122},
  bibsource    = {dblp computer science bibliography, https://dblp.org}
}

@inproceedings{DBLP:conf/cvpr/0001ZNMSVSTW025,
  author       = {Chaoyang Wang and
                  Peiye Zhuang and
                  Tuan Duc Ngo and
                  Willi Menapace and
                  Aliaksandr Siarohin and
                  Michael Vasilkovsky and
                  Ivan Skorokhodov and
                  Sergey Tulyakov and
                  Peter Wonka and
                  Hsin{-}Ying Lee},
  title        = {4Real-Video: Learning Generalizable Photo-Realistic 4D Video Diffusion},
  booktitle    = CVPR,
  pages        = {17723--17732},

  year         = {2025},

  doi          = {10.1109/CVPR52734.2025.01651},
  bibsource    = {dblp computer science bibliography, https://dblp.org}
}

@inproceedings{DBLP:conf/cvpr/LiuYL0Z25,
  author       = {Zhuoman Liu and
                  Weicai Ye and
                  Yan Luximon and
                  Pengfei Wan and
                  Di Zhang},
  title        = {Unleashing the Potential of Multi-modal Foundation Models and Video
                  Diffusion for 4D Dynamic Physical Scene Simulation},
  booktitle    = CVPR,
  pages        = {11016--11025},

  year         = {2025},

  doi          = {10.1109/CVPR52734.2025.01029},
  bibsource    = {dblp computer science bibliography, https://dblp.org}
}

@inproceedings{DBLP:conf/cvpr/PengZWXXZKTZ25,
  author       = {Chensheng Peng and
                  Chengwei Zhang and
                  Yixiao Wang and
                  Chenfeng Xu and
                  Yichen Xie and
                  Wenzhao Zheng and
                  Kurt Keutzer and
                  Masayoshi Tomizuka and
                  Wei Zhan},
  title        = {DeSiRe-GS: 4D Street Gaussians for Static-Dynamic Decomposition and
                  Surface Reconstruction for Urban Driving Scenes},
  booktitle    = CVPR,
  pages        = {6782--6791},

  year         = {2025},

  doi          = {10.1109/CVPR52734.2025.00636},
  bibsource    = {dblp computer science bibliography, https://dblp.org}
}

@inproceedings{DBLP:conf/cvpr/WuLHQZD25,
  author       = {Diankun Wu and
                  Fangfu Liu and
                  Yi{-}Hsin Hung and
                  Yue Qian and
                  Xiaohang Zhan and
                  Yueqi Duan},
  title        = {4D-Fly: Fast 4D Reconstruction from a Single Monocular Video},
  booktitle    = CVPR,
  pages        = {16663--16673},

  year         = {2025},

  doi          = {10.1109/CVPR52734.2025.01553},
  bibsource    = {dblp computer science bibliography, https://dblp.org}
}

@inproceedings{DBLP:conf/cvpr/AshutoshPG25,
  author       = {Kumar Ashutosh and
                  Georgios Pavlakos and
                  Kristen Grauman},
  title        = {FIction: 4D Future Interaction Prediction from Video},
  booktitle    = CVPR,
  pages        = {17613--17625},

  year         = {2025},
  doi          = {10.1109/CVPR52734.2025.01641},
  bibsource    = {dblp computer science bibliography, https://dblp.org}
}

@inproceedings{DBLP:conf/cvpr/SangCCMBC25,
  author       = {Lu Sang and
                  Zehranaz Canfes and
                  Dongliang Cao and
                  Riccardo Marin and
                  Florian Bernard and
                  Daniel Cremers},
  title        = {4Deform: Neural Surface Deformation for Robust Shape Interpolation},
  booktitle    = CVPR,
  pages        = {6542--6551},

  year         = {2025},

  doi          = {10.1109/CVPR52734.2025.00613},
  bibsource    = {dblp computer science bibliography, https://dblp.org}
}

@inproceedings{DBLP:conf/cvpr/MatsukiBD25,
  author       = {Hidenobu Matsuki and
                  Gwangbin Bae and
                  Andrew J. Davison},
  title        = {4DTAM: Non-Rigid Tracking and Mapping via Dynamic Surface Gaussians},
  booktitle    = CVPR,
  pages        = {26921--26932},

  year         = {2025},

  doi          = {10.1109/CVPR52734.2025.02507},
  bibsource    = {dblp computer science bibliography, https://dblp.org}
}

@inproceedings{DBLP:conf/cvpr/YaoZW25,
  author       = {David Yifan Yao and
                  Albert J. Zhai and
                  Shenlong Wang},
  title        = {Uni4D: Unifying Visual Foundation Models for 4D Modeling from a Single
                  Video},
  booktitle    = CVPR,
  pages        = {1116--1126},

  year         = {2025},

  doi          = {10.1109/CVPR52734.2025.00112},
  bibsource    = {dblp computer science bibliography, https://dblp.org}
}

@inproceedings{wu2025cat4d,
  title={Cat4d: Create anything in 4d with multi-view video diffusion models},
  author={Wu, Rundi and Gao, Ruiqi and Poole, Ben and Trevithick, Alex and Zheng, Changxi and Barron, Jonathan T and Holynski, Aleksander},
  booktitle=CVPR,
  pages={26057--26068},
  year={2025}
}

@inproceedings{yu2025dyn,
  title={Dyn-hamr: Recovering 4d interacting hand motion from a dynamic camera},
  author={Yu, Zhengdi and Zafeiriou, Stefanos and Birdal, Tolga},
  booktitle=CVPR,
  pages={27716--27726},
  year={2025}
}

@inproceedings{chu2025robust,
  title={Robust multi-object 4d generation for in-the-wild videos},
  author={Chu, Wen-Hsuan and Ke, Lei and Liu, Jianmeng and Huo, Mingxiao and Tokmakov, Pavel and Fragkiadaki, Katerina},
  booktitle=CVPR,
  pages={22067--22077},
  year={2025}
}

@article{DBLP:journals/corr/abs-2504-13152,
  author       = {Haiwen Feng and
                  Junyi Zhang and
                  Qianqian Wang and
                  Yufei Ye and
                  Pengcheng Yu and
                  Michael J. Black and
                  Trevor Darrell and
                  Angjoo Kanazawa},
  title        = {St4RTrack: Simultaneous 4D Reconstruction and Tracking in the World},
  journal      = {CoRR},
  volume       = {abs/2504.13152},
  year         = {2025},

  doi          = {10.48550/ARXIV.2504.13152},
  eprinttype    = {arXiv},
  eprint       = {2504.13152},
  bibsource    = {dblp computer science bibliography, https://dblp.org}
}

@inproceedings{yuan2025self,
  title={Self-Supervised Monocular 4D Scene Reconstruction for Egocentric Videos},
  author={Yuan, Chengbo and Chen, Geng and Yi, Li and Gao, Yang},
  booktitle=ICCV,
  pages={8863--8874},
  year={2025}
}

@article{lu2025humoto,
  title={HUMOTO: A 4D Dataset of Mocap Human Object Interactions},
  author={Lu, Jiaxin and Huang, Chun-Hao Paul and Bhattacharya, Uttaran and Huang, Qixing and Zhou, Yi},
  journal={arXiv:2504.10414},
  year={2025}
}

@article{DBLP:journals/corr/abs-2503-16396,
  author       = {Chun{-}Han Yao and
                  Yiming Xie and
                  Vikram Voleti and
                  Huaizu Jiang and
                  Varun Jampani},
  title        = {{SV4D} 2.0: Enhancing Spatio-Temporal Consistency in Multi-View Video
                  Diffusion for High-Quality 4D Generation},
  journal      = {CoRR},
  volume       = {abs/2503.16396},
  year         = {2025},

  doi          = {10.48550/ARXIV.2503.16396},
  eprinttype    = {arXiv},
  eprint       = {2503.16396},
  bibsource    = {dblp computer science bibliography, https://dblp.org}
}

@article{DBLP:journals/corr/abs-2504-07961,
  author       = {Zeren Jiang and
                  Chuanxia Zheng and
                  Iro Laina and
                  Diane Larlus and
                  Andrea Vedaldi},
  title        = {Geo4D: Leveraging Video Generators for Geometric 4D Scene Reconstruction},
  journal      = {CoRR},
  volume       = {abs/2504.07961},
  year         = {2025},

  doi          = {10.48550/ARXIV.2504.07961},
  eprinttype    = {arXiv},
  eprint       = {2504.07961},
  bibsource    = {dblp computer science bibliography, https://dblp.org}
}

@article{DBLP:journals/corr/abs-2502-08377,
  author       = {Liying Yang and
                  Chen Liu and
                  Zhenwei Zhu and
                  Ajian Liu and
                  Hui Ma and
                  Jian Nong and
                  Yanyan Liang},
  title        = {Not All Frame Features Are Equal: Video-to-4D Generation via Decoupling
                  Dynamic-Static Features},
  journal      = {CoRR},
  volume       = {abs/2502.08377},
  year         = {2025},

  doi          = {10.48550/ARXIV.2502.08377},
  eprinttype    = {arXiv},
  eprint       = {2502.08377},
  bibsource    = {dblp computer science bibliography, https://dblp.org}
}

@article{DBLP:journals/corr/abs-2507-23782,
  author       = {Zihan Wang and
                  Jeff Tan and
                  Tarasha Khurana and
                  Neehar Peri and
                  Deva Ramanan},
  title        = {MonoFusion: Sparse-View 4D Reconstruction via Monocular Fusion},
  journal      = {CoRR},
  volume       = {abs/2507.23782},
  year         = {2025},

  doi          = {10.48550/ARXIV.2507.23782},
  eprinttype    = {arXiv},
  eprint       = {2507.23782},
  bibsource    = {dblp computer science bibliography, https://dblp.org}
}

@article{DBLP:journals/corr/abs-2507-23785,
  author       = {Bowen Zhang and
                  Sicheng Xu and
                  Chuxin Wang and
                  Jiaolong Yang and
                  Feng Zhao and
                  Dong Chen and
                  Baining Guo},
  title        = {Gaussian Variation Field Diffusion for High-fidelity Video-to-4D Synthesis},
  journal      = {CoRR},
  volume       = {abs/2507.23785},
  year         = {2025},

  doi          = {10.48550/ARXIV.2507.23785},
  eprinttype    = {arXiv},
  eprint       = {2507.23785},
  bibsource    = {dblp computer science bibliography, https://dblp.org}
}

@article{DBLP:journals/corr/abs-2503-09631,
  author       = {Jianqi Chen and
                  Biao Zhang and
                  Xiangjun Tang and
                  Peter Wonka},
  title        = {{V2M4:} 4D Mesh Animation Reconstruction from a Single Monocular Video},
  journal      = {CoRR},
  volume       = {abs/2503.09631},
  year         = {2025},

  doi          = {10.48550/ARXIV.2503.09631},
  eprinttype    = {arXiv},
  eprint       = {2503.09631},
  bibsource    = {dblp computer science bibliography, https://dblp.org}
}

@inproceedings{DBLP:conf/eccv/MihajlovicPTMBTB24,
  author       = {Marko Mihajlovic and
                  Sergey Prokudin and
                  Siyu Tang and
                  Robert Maier and
                  Federica Bogo and
                  Tony Tung and
                  Edmond Boyer},
  editor       = {Ales Leonardis and
                  Elisa Ricci and
                  Stefan Roth and
                  Olga Russakovsky and
                  Torsten Sattler and
                  G{\"{u}}l Varol},
  title        = {SplatFields: Neural Gaussian Splats for Sparse 3D and 4D Reconstruction},
  booktitle    = ECCV,
  series       = {Lecture Notes in Computer Science},
  volume       = {15060},
  pages        = {313--332},

  year         = {2024},

  doi          = {10.1007/978-3-031-72627-9\_18},
  bibsource    = {dblp computer science bibliography, https://dblp.org}
}

@inproceedings{DBLP:conf/eccv/WuYJCWB24,
  author       = {Zijie Wu and
                  Chaohui Yu and
                  Yanqin Jiang and
                  Chenjie Cao and
                  Fan Wang and
                  Xiang Bai},
  editor       = {Ales Leonardis and
                  Elisa Ricci and
                  Stefan Roth and
                  Olga Russakovsky and
                  Torsten Sattler and
                  G{\"{u}}l Varol},
  title        = {{SC4D:} Sparse-Controlled Video-to-4D Generation and Motion Transfer},
  booktitle    = ECCV,
  series       = {Lecture Notes in Computer Science},
  volume       = {15071},
  pages        = {361--379},

  year         = {2024},

  doi          = {10.1007/978-3-031-72624-8\_21},
  bibsource    = {dblp computer science bibliography, https://dblp.org}
}

@inproceedings{DBLP:conf/eccv/BahmaniLYSRLLPTWTL24,
  author       = {Sherwin Bahmani and
                  Xian Liu and
                  Wang Yifan and
                  Ivan Skorokhodov and
                  Victor Rong and
                  Ziwei Liu and
                  Xihui Liu and
                  Jeong Joon Park and
                  Sergey Tulyakov and
                  Gordon Wetzstein and
                  Andrea Tagliasacchi and
                  David B. Lindell},
  editor       = {Ales Leonardis and
                  Elisa Ricci and
                  Stefan Roth and
                  Olga Russakovsky and
                  Torsten Sattler and
                  G{\"{u}}l Varol},
  title        = {{TC4D:} Trajectory-Conditioned Text-to-4D Generation},
  booktitle    = ECCV,
  series       = {Lecture Notes in Computer Science},
  volume       = {15104},
  pages        = {53--72},

  year         = {2024},

  doi          = {10.1007/978-3-031-72952-2\_4},
  bibsource    = {dblp computer science bibliography, https://dblp.org}
}

@article{sam3d,
      title={SAM 3D: 3Dfy Anything in Images}, 
      author={SAM 3D Team and Xingyu Chen and Fu-Jen Chu and Pierre Gleize and Kevin J Liang and Alexander Sax and Hao Tang and Weiyao Wang and Michelle Guo and Thibaut Hardin and Xiang Li and Aohan Lin and Jiawei Liu and Ziqi Ma and Anushka Sagar and Bowen Song and Xiaodong Wang and Jianing Yang and Bowen Zhang and Piotr Dollár and Georgia Gkioxari and Matt Feiszli and Jitendra Malik},
      year={2025},
      eprint={2511.16624},
      archivePrefix={arXiv},
      primaryClass={cs.CV},
 
}

@inproceedings{chen2026v,
  title={V-HOI: Velocity-Aware Human-Object Interaction Generation},
  author={Chen, Honghui and Zhou, Fan and Wang, Ruomei and Zhao, Baoquan},
  booktitle={International Conference on Multimedia Modeling},
  pages={519--532},
  year={2026},

}

@article{wang2026multimodal,
  title={Multimodal Priors-Augmented Text-Driven 3D Human-Object Interaction Generation},
  author={Wang, Yin and Zhang, Ziyao and Leng, Zhiying and Liu, Haitian and Li, Frederick WB and Li, Mu and Liang, Xiaohui},
  journal={arXiv:2602.10659},
  year={2026}
}

@article{liu2026open,
  title={Open-Vocabulary Functional 3D Human-Scene Interaction Generation},
  author={Liu, Jie and Sun, Yu and Cseke, Alpar and Feng, Yao and Heron, Nicolas and Black, Michael J and Zhang, Yan},
  journal={arXiv:2601.20835},
  year={2026}
}

@article{chen2026motion,
  title={Motion 3-to-4: 3D Motion Reconstruction for 4D Synthesis},
  author={Chen, Hongyuan and Chen, Xingyu and Zhang, Youjia and Xu, Zexiang and Chen, Anpei},
  journal={arXiv:2601.14253},
  year={2026}
}

@article{liang2026dextercap,
  title={DexterCap: An Affordable and Automated System for Capturing Dexterous Hand-Object Manipulation},
  author={Liang, Yutong and Xu, Shiyi and Zhang, Yulong and Zhan, Bowen and Zhang, He and Liu, Libin},
  journal={arXiv:2601.05844},
  year={2026}
}

@article{gupta2026pokenet,
  title={PokeNet: Learning Kinematic Models of Articulated Objects from Human Observations},
  author={Gupta, Anmol and Gu, Weiwei and Patil, Omkar and Lee, Jun Ki and Gopalan, Nakul},
  journal={arXiv:2602.02741},
  year={2026}
}

@article{kim2026camo,
  title={CAMO: Category-Agnostic 3D Motion Transfer from Monocular 2D Videos},
  author={Kim, Taeyeon and Na, Youngju and Lee, Jumin and Sung, Minhyuk and Yoon, Sung-Eui},
  journal={arXiv:2601.02716},
  year={2026}
}
